\pdfoutput=1

\documentclass[11pt]{article}

\usepackage{ACL2023}

\usepackage{times}
\usepackage{fix-cm}
\usepackage{latexsym}
\usepackage{booktabs}
\usepackage{array}
\usepackage{setspace}
\usepackage{enumitem}
\setlist{leftmargin=1mm}
\usepackage{multirow}
\usepackage[normalem]{ulem}
\useunder{\uline}{\ul}{}
\usepackage{amsmath}
\usepackage{colortbl}
\usepackage{xcolor}

\usepackage[most]{tcolorbox}
\usepackage{xcolor}
\usepackage{multirow}
\usepackage{wasysym}

\definecolor{promptbg}{RGB}{248, 249, 250}
\definecolor{examplebg}{RGB}{248, 249, 250}
\definecolor{White}{RGB}{251, 251, 251}

\newtcolorbox{zeroshot}{
    enhanced,
    colback=promptbg,
    colframe=black!80,
    boxrule=0.5pt,
    arc=2mm,
    title=Zero-Shot Prompt,
    attach boxed title to top left={yshift=-2mm, xshift=2mm},
    boxed title style={
        colback=white,  
        colframe=black!80  
    },
    fonttitle=\bfseries,
    coltitle=blue,
    breakable
}

\newtcolorbox{fewshot}{
    enhanced,
    colback=promptbg,
    colframe=black!80,
    boxrule=0.5pt,
    arc=2mm,
    title=Few-Shot Prompt,
    attach boxed title to top left={yshift=-2mm, xshift=2mm},
    boxed title style={
        colback=white,  
        colframe=black!80  
    },
    fonttitle=\bfseries,
    coltitle=blue,
    breakable
}

\newtcolorbox{example}{
    enhanced,
    colback=examplebg,
    colframe=blue!20,
    boxrule=0.5pt,
    left=2mm,
    right=2mm,
    top=2mm,
    bottom=2mm,
    breakable
}

\definecolor{darkred}{RGB}{255,150,150}     
\definecolor{lightred}{RGB}{255,200,200}    
\definecolor{darkyellow}{RGB}{255,230,150}  
\definecolor{lightyellow}{RGB}{255,255,200} 
\definecolor{lightgreen}{RGB}{220,255,220}  
\definecolor{darkgreen}{RGB}{180,255,180}   

\usepackage[T1]{fontenc}

\usepackage[utf8]{inputenc}

\usepackage{microtype}

\usepackage{inconsolata}
\usepackage{graphicx}

\usepackage{tikz}
\usetikzlibrary{shapes.geometric, arrows, positioning, calc}

\tcbset{
  aibox/.style={
    top=10pt,
    colback=white,
    colframe=black,
    colbacktitle=black,
    enhanced,
    center,
    attach boxed title to top left={yshift=-0.1in,xshift=0.15in},
    boxed title style={boxrule=0pt,colframe=white,},
  }
}
\newtcolorbox{AIbox}[2][]{aibox, title=#2,#1}

\newcommand{\name}{\textsc{RADAR}}

\title{\name: A Reasoning-Guided Attribution Framework for\\Explainable Visual Data Analysis}

\author{\textbf{Anku Rani}$^{1}$\thanks{\,\,\,A part of the work was done when the author was at Adobe Research} \quad \textbf{Aparna Garimella}$^{2}$ \quad \textbf{Apoorv Saxena}$^{2}$ \\ \textbf{Balaji Vasan Srinivasan}$^{2}$ \quad \textbf{Paul Pu Liang}$^{1}$ \\
$^{1}$ Massachusetts Institute of Technology \quad     
$^{2}$ Adobe Research \quad \\
\tt  ankurani@mit.edu
}

\begin{document}
\maketitle

\begin{abstract}

Data visualizations like charts are fundamental tools for quantitative analysis and decision-making across fields, requiring accurate interpretation and mathematical reasoning. The emergence of Multimodal Large Language Models (MLLMs) offers promising capabilities for automated visual data analysis, such as processing charts, answering questions, and generating summaries. 
However, they provide no visibility into which parts of the visual data informed their conclusions; this black-box nature poses significant challenges to real-world trust and adoption. In this paper, we take the first major step towards evaluating and enhancing the capabilities of MLLMs to attribute their reasoning process by highlighting the specific regions in charts and graphs that justify model answers.
To this end, we contribute \name, a semi-automatic approach to obtain a benchmark dataset comprising 17,819 diverse samples with charts, questions, reasoning steps, and attribution annotations. We also introduce a method that provides attribution for chart-based mathematical reasoning. Experimental results demonstrate that our reasoning-guided approach improves attribution accuracy by 15\% compared to baseline methods and enhanced attribution capabilities translate to stronger answer generation, achieving an average BERTScore of $\sim$ 0.90, indicating high alignment with ground truth responses. This advancement represents a significant step toward more interpretable and trustworthy chart analysis systems, enabling users to verify and understand model decisions through reasoning and attribution.

\end{abstract}

\section{Introduction}

Data visualizations, particularly bar charts and line charts, serve as fundamental tools for data representation and analysis across domains. These visualizations enable informed decision-making by presenting complex numerical information in an accessible format. However, extracting insights from charts often requires sophisticated mathematical reasoning like complex trend analysis and comparative calculations \cite{10.1145/3313831.3376467}. The emergence of Multimodal Large Language Models (MLLMs) offers promising capabilities for automated visual data analysis \citep{satpute2024can, srivastava2024evaluating, ahn2024large,gupta2024polymathchallengingmultimodalmathematical}, but they do so without providing any visibility into which parts of the chart informed their decision-making process \cite{wang2023scientific}. This poses problems for real-world trust and adoption, especially since charts and diagrams are often used in sensitive applications like business, medicine, education, and policy \cite{islam2024large}.

\begin{figure*}[t]
    \centering
    \includegraphics[width=0.9\textwidth]{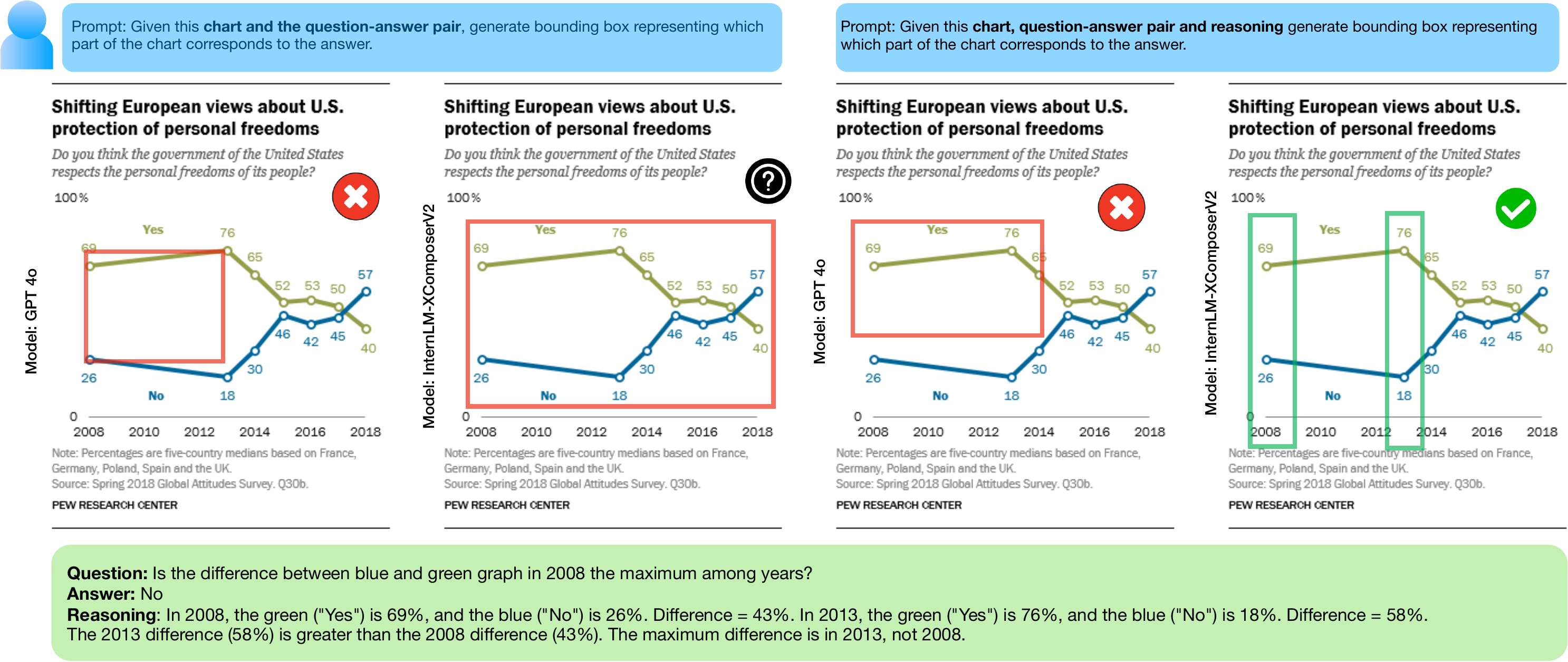}
    \caption{Comparison of attribution methods between GPT-4o \& \name\ (Our method that utilizes InternLM-XComposerV2). Left: Both models receive only the chart, question, and answer as input. Right: Models additionally receive reasoning steps, leading to more precise attributions. The example shows how incorporating reasoning steps helps InternLM-XComposer2 correctly attribute the relevant data points for comparing differences between lines in 2008 and 2013, while GPT-4o struggles with accurate attribution even with reasoning provided.}
    \label{fig:Intro_Img1}
\end{figure*}

In this paper, we take the first major step towards evaluating and enhancing the capabilities of MLLMs to attribute their reasoning process to visual data like charts and graphs. We introduce attribution to identify and highlight key regions within charts through bounding boxes that justify final decisions, helping make the reasoning process transparent and verifiable. Previous research on attribution has largely concentrated on text-based question-answering and general visual question-answering tasks \cite{yue2023automatic, phukan2024logitlenscontextualembeddings, phukan-etal-2024-peering, bohnet2022attributed, qi2024model}, which face significant limitations when applied to mathematical chart analysis. Figure \ref{fig:Intro_Img1} shows that existing methods struggle to accurately identify relevant chart regions that contribute to answers for complex mathematical questions. This gap is particularly notable as mathematical reasoning with charts requires precise identification of data points and understanding of their relationships for operations like aggregation, comparison, and trend analysis.

To bridge this gap, we introduce \textbf{\name}: A \textbf{R}e\textbf{A}soning-Guide\textbf{D} \textbf{A}tt\textbf{R}ibution Framework for Explainable Visual Data Analysis, a novel framework for attributing generated answers to specific regions in charts. \name\ has four main contributions:

\begin{enumerate}[noitemsep,topsep=0pt,nosep,leftmargin=*,parsep=0pt,partopsep=0pt]
    \item We introduce the \textbf{task of attribution for mathematical question answering in charts}, addressing a critical gap in current visual mathematical question answering, as outlined in Figure \ref{fig:Intro_Img1}.
    \item We present a systematic \textbf{data curation strategy} that combines MLLM-generated reasoning and attribution annotations with human corrections. This results in a high-quality dataset derived from ChartQA \cite{masry-etal-2022-chartqa}, comprising annotated examples spanning line and bar chart types and mathematical operations.
    \item We \textbf{propose an automatic attribution and reasoning generation method}
    for ChartQA and evaluate the generated attribution automatically by generating answers using feature attribution.
    \item We demonstrate that high-quality \textbf{reasoning steps substantially improve attribution accuracy}, achieving a 15\% average improvement over baseline methods, highlighting both the importance of reasoning in attribution tasks and identifying opportunities for further advances in reasoning generation.

    \item We establish that enhanced attribution capabilities lead to more accurate answer generation, as evidenced by high BERTScore values ($\sim$0.90) shown in Table \ref{tab:similarity_scores}, demonstrating the \textbf{synergistic relationship between attribution and answer quality}.
\end{enumerate}

\section{Related Work \& Attribution Definition}

Recent work has increasingly focused on attribution mechanisms for AI systems' outputs. For text systems, \citet{bohnet2022attributed} survey attribution methods in open-domain generation, while \citet{phukan-etal-2024-peering} and \citet{qi2024model} advance answer attribution through hidden state analysis and the MIRAGE approach respectively. For multimodal systems, \citet{phukan2024logitlenscontextualembeddings} extend logit lens techniques to detect visual hallucinations. Chart-based question answering has evolved significantly, with the ChartQA dataset \cite{masry-etal-2022-chartqa} providing a comprehensive benchmark and advances like Chart Llama \cite{han2023chartllama} and ChartOCR \cite{luo2021chartocr} improving chart understanding capabilities.

Mathematical reasoning has become crucial for chart interpretation, with \citet{imani2023mathprompter} introducing MathPrompter for multiple solution paths and \citet{ranaldi-freitas-2024-aligning} addressing Chain-of-Thought limitations. As \citet{lu-etal-2023-survey} note, mathematical reasoning serves as a key testbed for AI capabilities. While models like InternLM-XComposer2 \cite{dong2024internlm} excel in multimodal understanding, existing attribution methods face challenges with mathematical chart questions, creating a critical gap in trustworthy chart-based mathematical reasoning systems.

\subsection{Attribution Definition}

\begin{figure*}[!ht]
    \centering
    \includegraphics[width=1\textwidth]{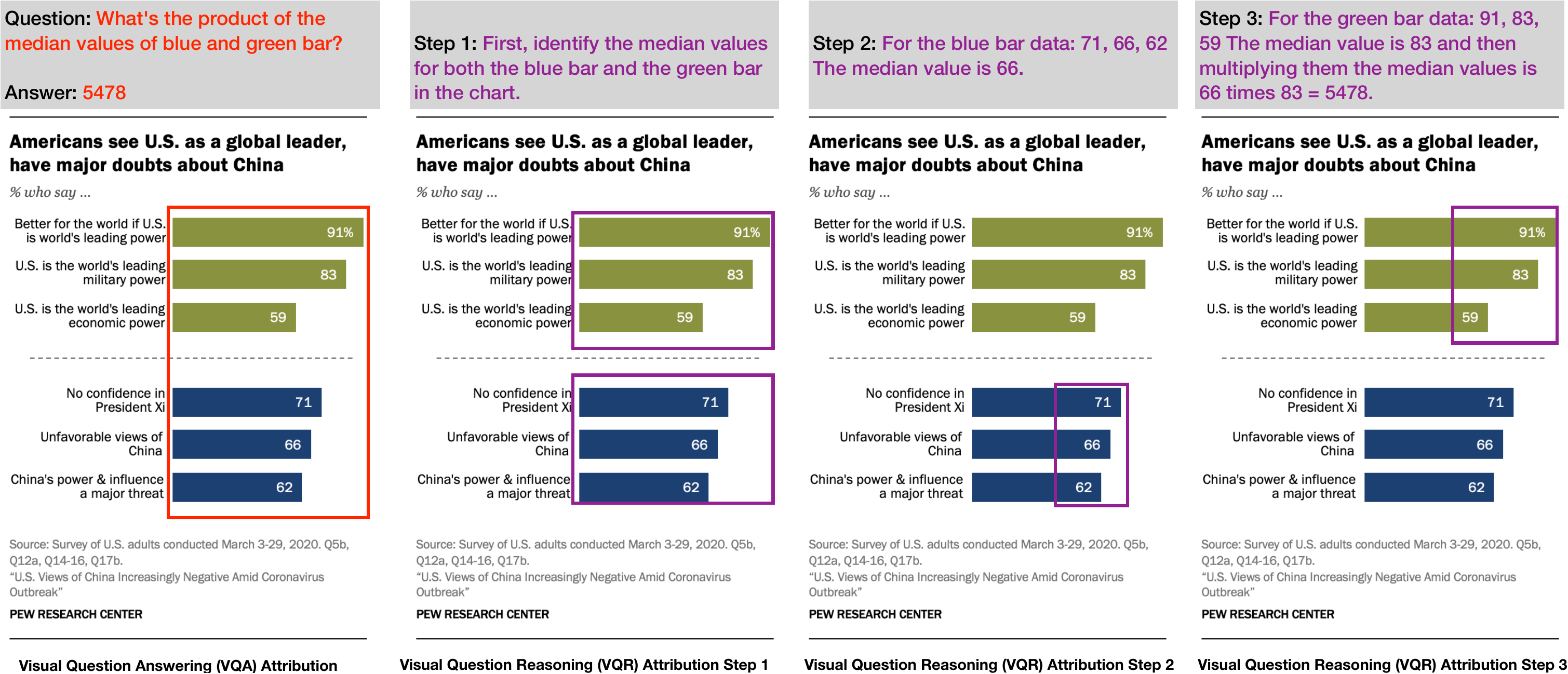}
    \caption{\name's attribution process for mathematical reasoning on charts. Left to right: (1) VQA attribution highlights all relevant data bars needed for computing the final answer, (2) Step 1 attribution identifies the specific bars needed to find median values, (3) Step 2 attribution focuses on the blue bars for calculating their median (66), and (4) Step 3 attribution shows the green bars used for the final multiplication step (66 × 83 = 5478). The attribution represents how our framework traces both the final answer \& intermediate reasoning steps.}
    \label{fig:attribution_steps}
\end{figure*}
Chart attribution aims to identify regions of a chart that support generated answers, similar to the Grounded Visual Question Answering (VQA) approach proposed by \cite{phukan2024logitlenscontextualembeddings}. For mathematical chart question answering, where complex reasoning steps are essential to arrive at answers, we propose a two-level attribution framework that provides transparency not only for final answers but also for intermediate reasoning steps.

\subsection{Answer-Level Attribution}
Basic chart attribution involves visually linking chart elements to answers using bounding boxes, highlighting the specific data points that support the answer. In the leftmost chart of Figure \ref{fig:attribution_steps}, while the bounding boxes highlight all data points contributing to the answer "5478", the reasoning behind this calculation remains unclear without additional context. This demonstrates why incorporating reasoning steps becomes crucial for questions involving mathematical operations, where the path to the answer is as important as the answer itself.

\subsection{Reasoning-Level Attribution}
When solving mathematical questions with charts, the path to the answer often involves multiple reasoning steps. Our framework attributes each reasoning step to relevant chart regions, creating a traceable connection between the reasoning process and visual elements. As shown in Figure \ref{fig:attribution_steps}, the 2nd, 3rd and 4th chart represents each of the reasoning steps. This granular-level attribution approach enhances trust in the system by making both final answers and the reasoning process transparent and verifiable against the source chart.

Additional examples for both answer level and reasoning level attribution are present in the Appendix section \ref{Appendix_Attribution_Definition}.

\vspace{-2mm}
\section{Dataset Curation}
Currently, no datasets exist that provide reasoning steps for chart question answering or attribution annotations for mathematical chart QA. To address this gap, we first examine existing model capabilities before developing a semi-automatic annotation strategy.

\subsection{Reasoning Capabilities of MLLMs for Charts}

Based on performance evaluations in the Polymath benchmark \cite{gupta2024polymathchallengingmultimodalmathematical}, we select Claude 3.5 Sonnet and GPT-4o as our primary models for analysis. We also include GPT-4v for its vision capabilities. To assess the performance on reasoning generation, we randomly select 100 examples from the ChartQA dataset \cite{masry-etal-2022-chartqa}, each containing a chart, question, and answer triple. Charts and questions are passed to these models as input and they are prompted to generate answers and reasoning. These answers and reasoning are annotated by human annotators on whether they are correct or not, and the results are presented in Table \ref{tab:benchmarking}.

\begin{table}[t]
\centering
\resizebox{\linewidth}{!}{%
\begin{tabular}{lcc}
\toprule
\textbf{Model} & \textbf{Answer is Correct} & \textbf{Reasoning is Correct} \\
& \textbf{(Human Annotated)} & \textbf{(Human Annotated)} \\
\midrule
\textbf{Gpt-4o} & \cellcolor{yellow!60!red}{58\%} & \cellcolor{yellow!40!red}{49\%} \\
\textbf{Gpt-4v} & \cellcolor{yellow!70!red}{64\%} & \cellcolor{yellow!35!red}{45\%} \\
\textbf{Claude-3.5-sonnet} & \cellcolor{green!85!yellow}{96\%} & \cellcolor{yellow!90!green}{75\%} \\
\bottomrule
\end{tabular}%
}
\caption{Benchmarking performance based on Visual Question Answering and Visual Question Reasoning.}
\label{tab:benchmarking}
\end{table}

As shown in Table \ref{tab:benchmarking}, while Claude 3.5 Sonnet demonstrates strong answer generation (96\% accuracy), its reasoning capabilities show significant room for improvement (75\% accuracy). Other models perform notably worse, with reasoning accuracies below 50\%.

A more detailed error analysis, including word clouds of failure patterns in appendix section \ref{data_sources} figure \ref{fig:wordcloud}, taxonomy in Appendix section \ref{data_sources} Figure \ref{fig:taxonomy} and specific examples is provided in the appendix section \ref{data_sources} Figure \ref{fig:reasoning_error}. Notably, providing correct answers alongside questions reduced reasoning failures from 51\% to 25\%, suggesting the potential for improved performance through better model guidance.

\subsection{Task Setup}
\label{Task_Setup}

Our goal is to obtain three types of annotations for chart-based mathematical attribution: (1) reasoning steps for given chart-question-answer triples, (2) answer attribution and (3) reasoning step attribution. 
Rather than annotating from scratch, we developed a semi-automatic approach leveraging Claude 3.5 Sonnet's capabilities to generate initial annotations for human correction.

We recruited two qualified annotators through the Upwork platform\footnote{\url{https://www.upwork.com}} after an initial screening of three candidates using 100 sample data points. The entire annotation process went for 120 hours and each annotator was paid 15 USD hourly.

For attribution annotation, we employed the VGG Image Annotator platform\footnote{\url{https://annotate.officialstatistics.org/}}, which provides an intuitive interface for drawing bounding boxes and mapping them to textual reasoning steps. Screenshots from the annotation interface, more details on initial screening and examples of such annotations are provided in the appendix section \ref{Appendix_Attribution_Definition} Figure \ref{fig:attribution_vqa} and \ref{fig:attribution_vqr}.

\textbf{Stage 1: Reasoning Validation and Correction.}
In Stage 1, annotators perform reasoning validation through three key steps: (1) correction by reviewing Claude-3.5-sonnet generated reasoning for chart-question-answer triples, (2) providing a binary correctness assessment (Yes/No) for each triple, and (3) categorizing errors in incorrect reasoning (such as color mismatches or illogical conclusions) while supplying corrected reasoning when the original is found to be inaccurate.

\textbf{Stage 2: Answer Attribution.}
For each chart-question-answer triple, annotators draw bounding boxes using the VGG image annotator indicating chart regions supporting the answer. Figure \ref{fig:attribution_vqa} demonstrates this process, showing both input and resulting annotations.

\textbf{Stage 3: Reasoning Attribution.}
Using the validated reasoning from Stage 1, annotators are instructed to provide bounding boxes using the VGG image annotator for each reasoning step. As shown in Figure \ref{fig:attribution_vqr}, each statement (e.g., "orange line represents unfavorable") is linked to relevant chart regions.

\subsection{Data Annotation \& Analysis}
To ensure annotation quality, we conducted initial screening to select mathematically proficient annotators, measured inter-annotator agreement using Kappa score \cite{cohen1960coefficient}, and had authors manually verify a sample of annotations. This semi-automatic approach significantly reduced annotation effort while maintaining high quality through human validation and correction. More details on the Inter-annotator Agreement consisting of Kappa score and Intersection Over Union (IOU) score \cite{rezatofighi2019generalized} calculation formula are present in the Appendix section \ref{Data_Annotation}.

\begin{table}[t]
\centering
\resizebox{\linewidth}{!}{%
\begin{tabular}{lccc}
\toprule
\textbf{Chart} & \textbf{Stage 1} & \textbf{Stage 2} & \textbf{Stage 3} \\
\textbf{Type} & \textbf{[Kappa Score]} & \textbf{[IOU Score]} & \textbf{[IOU Score]} \\
\midrule
\textbf{Line} & \cellcolor{green!85!yellow}{0.825} & \cellcolor{yellow!70!red}{0.524} & \cellcolor{yellow!60!red}{0.561} \\
\textbf{Bar} & \cellcolor{green!85!yellow}{0.920} & \cellcolor{yellow!60!red}{0.579} & \cellcolor{yellow!70!red}{0.647} \\
\bottomrule
\end{tabular}%
}
\caption{Across three annotation stages, we observed high agreement for reasoning validation (Kappa > 0.8), moderate agreement for answer-based attribution (IOU ~0.5), and improved agreement with reasoning-based attribution (IOU 0.56-0.64), indicating reasoning steps enhance annotation consistency.}
\label{tab:annotation_scores}
\end{table}

\subsection{Data Analysis}
\label{curated Data}

\begin{figure*}[!ht] 
    \centering
    \includegraphics[width=1\textwidth]{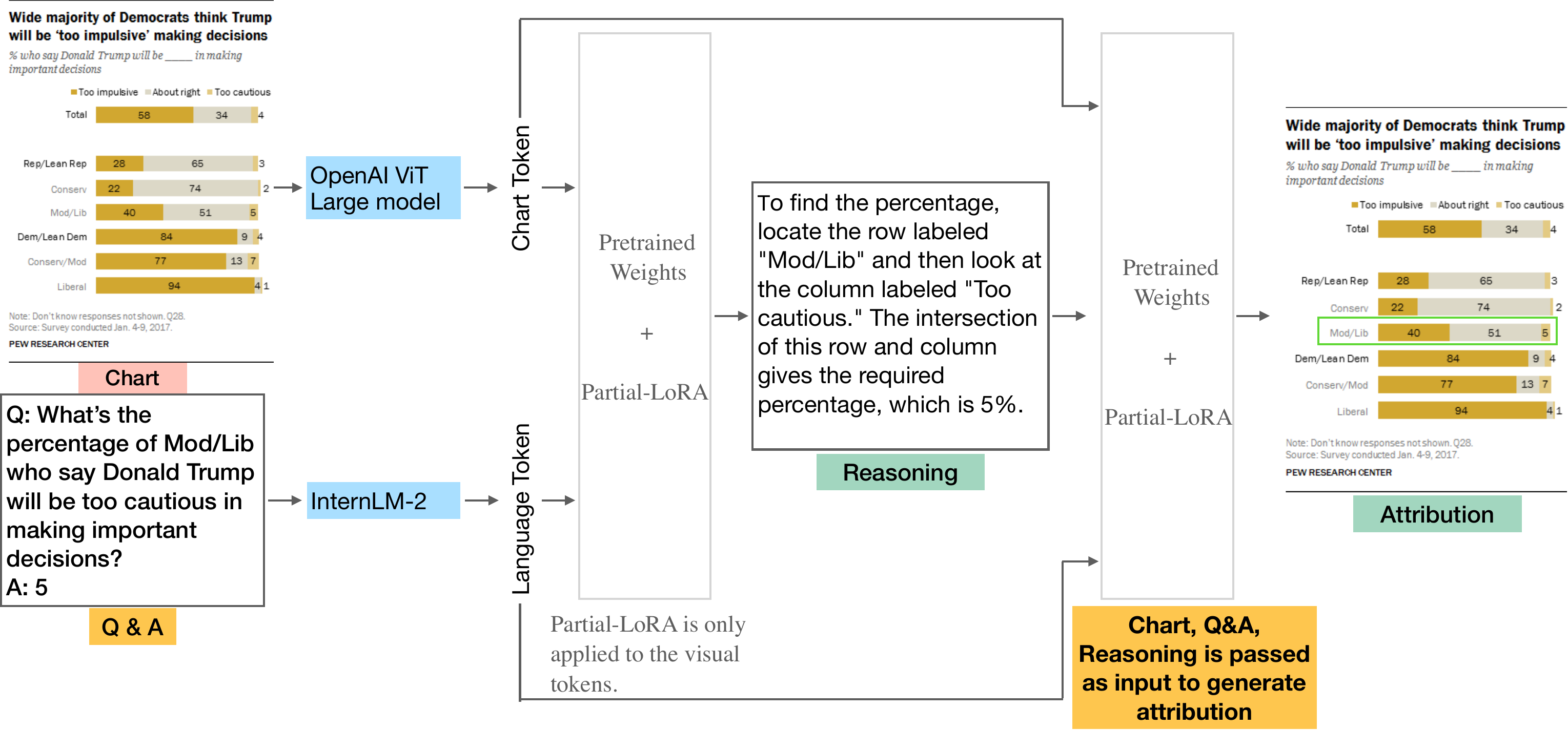}
    \caption{Overview of our proposed \name\ framework for reasoning and attribution generation using InternLM-XComposer2. Given a chart and Q\&A pair, the model processes visual tokens through CLIP ViT-Large and applies Partial-LoRA for chart-specific feature adaptation, while InternLM-2 processes textual inputs. The final output provides both answers and reasoning. This process is repeated by passing Chart, Q\&A pair, and generated reasoning to InternLM-XComposer2 to generate attribution.}
    \label{fig:method_reasoning}
\end{figure*}

After the annotation process, the key statistics about the data are summarized in the table \ref{tab:chart-data}. Table \ref{tab:chart-data} shows the breakdown of the dataset by chart type. There are a total of 1000 charts, consisting of 500 line charts and 500 bar charts. For each chart, there are 2 QA pairs, resulting in a total of 2000 QA pairs. Additionally, the annotators identified a total of 3599 reasoning steps across all the charts (stage 1).
The table also shows the number of image regions that were attributed to the QA-based annotations (stage 2) and the reasoning-based annotations (stage 3). For line charts, there are 1465 QA-based and 2691 reasoning-based attributed regions, while for bar charts, there are 2627 QA-based and 4437 reasoning-based attributed regions. The total number of data points is 17,819 (1000+2000+3599+4092+7128).

\begin{table}[t]
\centering
\resizebox{0.49\textwidth}{!}{%
\begin{tabular}{@{}l r r r r r@{}}
\toprule
Chart & Charts & QA & Reasoning & Attributed & Attributed \\
Type &  & Pairs & Steps & Regions (QA) & Regions (Reasoning) \\
\midrule
Line & 500 & 1000 & 1773 & 1465 & 2691 \\
Bar & 500 & 1000 & 1826 & 2627 & 4437 \\
\midrule
\textbf{Total} & \textbf{1000} & \textbf{2000} & \textbf{3599} & \textbf{4092} & \textbf{7128} \\
\bottomrule
\end{tabular}%
}
\caption{\name\ Dataset: 500 each of line and bar charts with 2 QA pairs per chart. The dataset progresses from QA pairs through reasoning steps to attributed regions, with reasoning-based attribution requiring 7128 regions versus 4092 for QA-based attribution.}
\label{tab:chart-data}
\end{table}

\section{\name: Proposed Method}

Our proposed method, \name, addresses the challenge of attributing mathematical reasoning in charts through a two-stage pipeline. Given a chart-question-answer triple, we first generate step-by-step reasoning using InternLM-XComposer2 model (figure \ref{fig:method_reasoning}), then leverage these reasoning steps along with the chart, question, and answer to produce attribution bounding boxes for both the final answer and intermediate reasoning steps. For reasoning generation, we utilize InternLM-XComposer2\footnote{\url{https://github.com/InternLM/InternLM-XComposer}} output. The model architecture incorporates a vision encoder (CLIP ViT-Large \cite{Radford2021LearningTV}) that processes charts into a 35×35 grid (1225 visual tokens) and maps them to a shared 4096-dimensional embedding space with text from InternLM-2 \cite{cai2024internlm2technicalreport}. Using \cite{dong2024internlm}, we employ Partial LoRA \cite{hu2021lora} (PLoRA), which applies additional trainable parameters specifically to visual tokens while preserving the base 7B-parameter language model's capabilities.
 
We utilize \cite{phukan2024logitlenscontextualembeddings}'s findings on attribution, and extract hidden states from layer 16, which empirically provides optimal semantic representations for our task. The attribution mechanism employs a GPU-accelerated sliding window approach, efficiently processing window configurations from 3×3 to 35×35 patches through normalized patch embedding averaging and cosine similarity metrics between textual descriptions and visual regions.

\section{Experiments}

We conduct experiments on the curated dataset presented in section \ref{curated Data}. We experiment on this dataset for three tasks i.e. (i) Attribution based on Visual Question Answering (VQA), (ii) Attribution based on Visual Question Reasoning (VQR), and (iii) Answer generation based on feature attribution.

\subsection{Baselines}
We evaluate \name\ against three state-of-the-art MLLMs: GPT-4o, GPT-4v, and Claude 3.5 Sonnet. For each baseline, we test both zero-shot and few-shot prompting strategies for two tasks: attribution based on answer attribution and reasoning attribution. We evaluate the generated attribution using answer generation using feature attribution.
\\
\noindent \textbf{Answer Attribution (VQA):} Models must identify relevant chart regions using bounding boxes that support their answers to specific questions. 
\\
\noindent \textbf{Reasoning Attribution (VQR):} Models must attribute their mathematical reasoning steps using bounding boxes to specific chart elements. Each reasoning step has different granular attribution as described in VQR steps 1, 2, and 3 of fig \ref{fig:attribution_steps}.
\\
We collect attribution results through API calls to GPT-4o\footnote{GPT-4o: "GPT-4o", "2023-05-15"}, GPT-4v\footnote{GPT-4v: "gpt-4-vision-preview", "2023-07-01-preview"}, and Claude 3.5 Sonnet\footnote{Claude 3.5 Sonnet: "claude-3-5-sonnet-20240620-v1:0"}. Since these models cannot directly output chart images with bounding boxes, we design prompts to obtain coordinates of the bounding boxes in the format of X1, Y1, X2, and Y2.

For \textbf{Answer Attribution (VQA)}, we explored two prompting strategies for answer attribution: (1) Zero-shot, where we used a base prompt without examples, and (2) Few-shot, where we included example cases in the prompt. Both prompting approaches are detailed in Appendix section \ref{Appendix:prompts}.

For \textbf{Reasoning Attribution (VQR)}, we conducted both zero-shot and few-shot prompting for our visual question reasoning attribution. The detailed prompts are available in Appendix section \ref{experiments}, with zero-shot prompts shown in figure \ref{fig:attribution_VQR_prompt_zeroshot} and few-shot prompts in figure \ref{fig:attribution_VQR_prompt_fewshot}. Input charts were base64 encoded for prompt delivery and decoded after collecting bounding box coordinates for subsequent analysis.

\subsection{\name-Automatic Reasoning Step Generation}

For automatic reasoning step generation, we leverage the Partial LoRA framework from \cite{dong2024internlm} due to its effectiveness in preserving language capabilities while adapting to visual inputs. Given a chart-question-answer triple $(C, Q, A)$, our goal is to generate reasoning steps $R$ that explain the answer derivation while maintaining alignment with visual elements.

Following the existing Partial LoRA architecture, we process inputs $x = [x_v, x_t]$, where $x_v$ represents visual tokens from the chart processed through CLIP ViT-Large, and $x_t$ represents the concatenated question-answer tokens. The output features are computed as follows:
\begin{equation}
    \hat{x} = [\hat{x}_v, \hat{x}_t]
\end{equation}
where $\hat{x}_t$ follows the standard language model path, and $\hat{x}_v$ incorporates visual adaptation through the Partial LoRA matrices.

A reasoning generator is used to maximize:
\begin{equation}
    P(R|C,Q,A) = \prod_{i=1}^{n} P(r_i|r_{<i},C,Q,A)
\end{equation}
where $r_i$ represents the $i$-th token in the reasoning sequence.

This approach enables our model to generate step-by-step reasoning by utilizing chart-specific visual features while maintaining strong language capabilities, producing coherent explanations that explicitly reference chart elements, and describing the mathematical operations needed to arrive at the answer. The generated reasoning provides a transparent explanation of the answer derivation process, which is then used to guide our attribution mechanism for identifying relevant chart regions.

\subsection{\name-Answer Generation Using Feature Attribution}
\label{new method}

To automatically evaluate \name\ without involving humans, we implemented a feature attribution mechanism that emphasizes the salient regions of the chart. Specifically, we identify the bounding box containing the ground truth answer and mask the remainder of the image by setting all pixels outside this region to zero. This preprocessed chart, along with the original question, is then passed to \name\ for answer extraction as described in Appendix section \ref{Appendix:Reasoning_Evaluation} Figure \ref{fig:Reverse_Evaluation}. We evaluate the effectiveness of this approach by computing a similarity score between the model's extracted answer and the ground truth which leads to automatic evaluation of \name.

\subsection{Metrics}

\begin{table}[t]
\centering
\resizebox{1\linewidth}{!}{
\Large  
\begin{tabular}{lccccc}
\toprule
\textbf{Chart} & & \textbf{BERTScore} & & \textbf{Semantic Textual} \\
\textbf{Type} & Avg & Avg & Avg & \textbf{Similarity} \\
 & Precision & Recall & F1 & \\
\midrule
\textbf{Line} & \cellcolor{green!80}{0.8928} & \cellcolor{green!70}{0.8855} & \cellcolor{green!75}{0.8890} & \cellcolor{yellow!85!red}{0.7383} \\
\textbf{Bar} & \cellcolor{green!80}{0.8928} & \cellcolor{green!70}{0.8866} & \cellcolor{green!75}{0.8895} & \cellcolor{yellow!80!red}{0.7428} \\
\bottomrule
\end{tabular}%
}
\vspace{-2mm}
\caption{Comparison of reasoning quality between \name-generated and human-annotated explanations. The evaluation uses two metrics: BERTScore and Semantic Textual Similarity (STS). Both chart types show consistent performance with BERTScore values around 0.89 and STS scores above 0.73, indicating strong alignment between machine-generated and human reasoning}
\label{tab:reasoning_scores}
\end{table}

\textbf{Reasoning Task:} In figure \ref{fig:method_reasoning}, automatic reasoning is generated by our proposed method \name. We also have human-annotated reasoning collected from stage 1 of Task setup as described in section \ref{Task_Setup}. We evaluate the generated reasoning using BERTScore \cite{Zhang2019BERTScoreET} \& Semantic Textual Similarity(STS) \cite{agirre-etal-2012-semeval} report the scores in Table \ref{tab:reasoning_scores}. Additional details are present in Appendix section \ref{Appendix:Reasoning_Evaluation}.

\noindent \textbf{Attribution Task:} 
We use \noindent \textbf{Multiple Box IOU Score} by extending the IOU score \cite{rezatofighi2019generalized} from single to multiple bounding boxes representing the utilization of pixels. While traditional single-box IOU simply computes the overlap between two rectangular regions, multiple-box IOU accounts for complex spatial relationships between sets of boxes, including internal overlaps within each set. Our implementation solves this challenge by converting box sets into binary masks, converting a geometric problem into a pixel-wise operation. This mask-based approach automatically handles cases where boxes within a set overlap, eliminating the need for complex geometric calculations of polygon intersections. The result is a Multi-Box IOU metric that maintains the intuitive interpretation of IOU while handling multiple boxes, helping to predict multiple regions for a single object or class.
\begin{equation}
IOU_{multi} = \frac{|\mathcal{M}_p \cap \mathcal{M}_{gt}|}{|\mathcal{M}_p \cup \mathcal{M}_{gt}|}
\end{equation}

where $\mathcal{M}p$ and $\mathcal{M}{gt}$ are binary masks generated from the sets of predicted and ground truth bounding boxes respectively.

\begin{table}[t]
\centering
\resizebox{0.48\textwidth}{!}{%
\begin{tabular}{l|cc|cc}
\toprule
\textbf{Model} & \multicolumn{2}{c|}{\textbf{VQA $\text{IOU}_{\text{multi}}$}} & \multicolumn{2}{c}{\textbf{VQR $\text{IOU}_{\text{multi}}$}} \\
& \textbf{Line} & \textbf{Bar} & \textbf{Line} & \textbf{Bar} \\
\midrule
GPT-4o (zero-shot) & \cellcolor{lightred}0.026 & \cellcolor{lightred}0.028 & \cellcolor{lightred}0.025 & \cellcolor{lightred}0.021 \\
GPT-4o (few-shot) & \cellcolor{lightred}0.020 & \cellcolor{lightred}0.022 & \cellcolor{lightred}0.022 & \cellcolor{darkred}0.019 \\
GPT-4v (zero-shot) & \cellcolor{darkred}0.016 & \cellcolor{darkred}0.019 & \cellcolor{lightred}0.021 & \cellcolor{lightred}0.023 \\
GPT-4v (few-shot) & \cellcolor{darkred}0.014 & \cellcolor{darkred}0.017 & \cellcolor{lightred}0.022 & \cellcolor{lightred}0.024 \\
Claude 3.5 (zero-shot) & \cellcolor{lightred}0.024 & \cellcolor{lightred}0.025 & \cellcolor{darkyellow}0.032 & \cellcolor{darkyellow}0.035 \\
Claude 3.5 (few-shot) & \cellcolor{lightred}0.025 & \cellcolor{lightred}0.021 & \cellcolor{darkyellow}0.037 & \cellcolor{darkyellow}0.039 \\
\midrule
RADAR (Automated Reasoning) & \cellcolor{lightgreen}\textbf{0.157} & \cellcolor{lightgreen}\textbf{0.153} & \cellcolor{lightgreen}\textbf{0.122} & \cellcolor{lightyellow}\textbf{0.082} \\
\midrule
RADAR (Human Reasoning) & \cellcolor{lightgreen}\textbf{0.157} & \cellcolor{lightgreen}\textbf{0.153} & \cellcolor{lightgreen}\textbf{0.136} & \cellcolor{lightgreen}\textbf{0.197} \\
\bottomrule
\end{tabular}%
}
\caption{Attribution performance comparison across different models and settings. Performance is measured using $IOU_{multi}$ scores for both VQA and VQR tasks.}
\label{tab:main_results}
\end{table}

\noindent \textbf{Answer Generation Task}: Attribution generated by \name\ as described in Figure \ref{fig:method_overall} is passed through the method described in section \ref{new method} to generate answers. We evaluate the generated answer using BERTScore \cite{Zhang2019BERTScoreET} \& Semantic Textual Similarity (STS) \cite{agirre-etal-2012-semeval} with human-annotated ones and report the scores in Table \ref{tab:similarity_scores}.
\vspace{-2mm}
\section{Results and Discussion}
\vspace{-2mm}
\noindent \textbf{Reasoning Task:} 
We report average BERTScore \& STS in table \ref{tab:reasoning_scores} between \name-generated reasoning and human-annotated reasoning. Results show strong alignment across chart types (precision: 0.8928, recall/F1: $\sim$0.89) with comparable STS scores ($\sim$0.74), demonstrating reliable reasoning capabilities that mirror human annotations.

\noindent \textbf{Attribution Task:}
Table \ref{tab:main_results} presents the Multi Box IOU scores across different models and prompting strategies. \name\ demonstrates substantial improvements: $504\%$ ($0.026\rightarrow0.157$) and $446\%$ ($0.028\rightarrow0.153$) for VQA tasks in line and bar charts respectively, while showing $230\%$ ($0.122\rightarrow0.037$) and $110\%$ ($0.039\rightarrow0.082$) improvement for VQR tasks with automated reasoning. The improvements increase further with human-validated reasoning, particularly for VQR tasks ($268\%$ for line charts, $405\%$ for bar charts). While these results demonstrate significant progress, the best-performing variant achieves IOU scores of $0.15-0.2$, indicating room for improvement through better attribution systems and MLLMs.

\noindent{\textbf{Answer Generation Task:}}
We take average STS and BERTScore to compare generated answers using feature attribution and the ground truth.
Table \ref{tab:similarity_scores} shows the effectiveness of feature attribution in \name\ for answer generation across line and bar charts. Bar charts consistently outperform line charts across all metrics, with higher Bertscore values in precision (0.9036), recall (0.8612), and F1 (0.8813). The moderate STS scores ($\sim$0.51) and marginal performance differences between chart types suggest \name's feature attribution approach is robust across visualization formats.

\begin{table}[t]
\centering
\resizebox{1\linewidth}{!}{%
\Large
\begin{tabular}{lccccc}
\toprule
\textbf{Chart} & & \textbf{Bertscore} & & \textbf{Semantic Textual} \\
\textbf{Type} & Avg & Avg & Avg & \textbf{Similarity} \\
& Precision & Recall & F1 & \\
\midrule
\textbf{Line} & \cellcolor{green!80}{0.8932} & \cellcolor{green!80}{0.8432} & \cellcolor{green!80}{0.8668} & \cellcolor{yellow!80}{0.5059} \\
\textbf{Bar} & \cellcolor{green!80}{0.9036} & \cellcolor{green!80}{0.8612} & \cellcolor{green!80}{0.8813} & \cellcolor{yellow!80}{0.5198} \\
\bottomrule
\end{tabular}%
}
\vspace{-2mm}
\caption{Comparison of textual similarity between \name-generated answers using feature attribution and ground truth using Bertscore and STS metrics across line and bar charts, demonstrating consistent performance across visualization formats.}
\label{tab:similarity_scores}
\end{table}

\begin{figure}[!t] 
    \centering
    \includegraphics[width=0.5\textwidth]{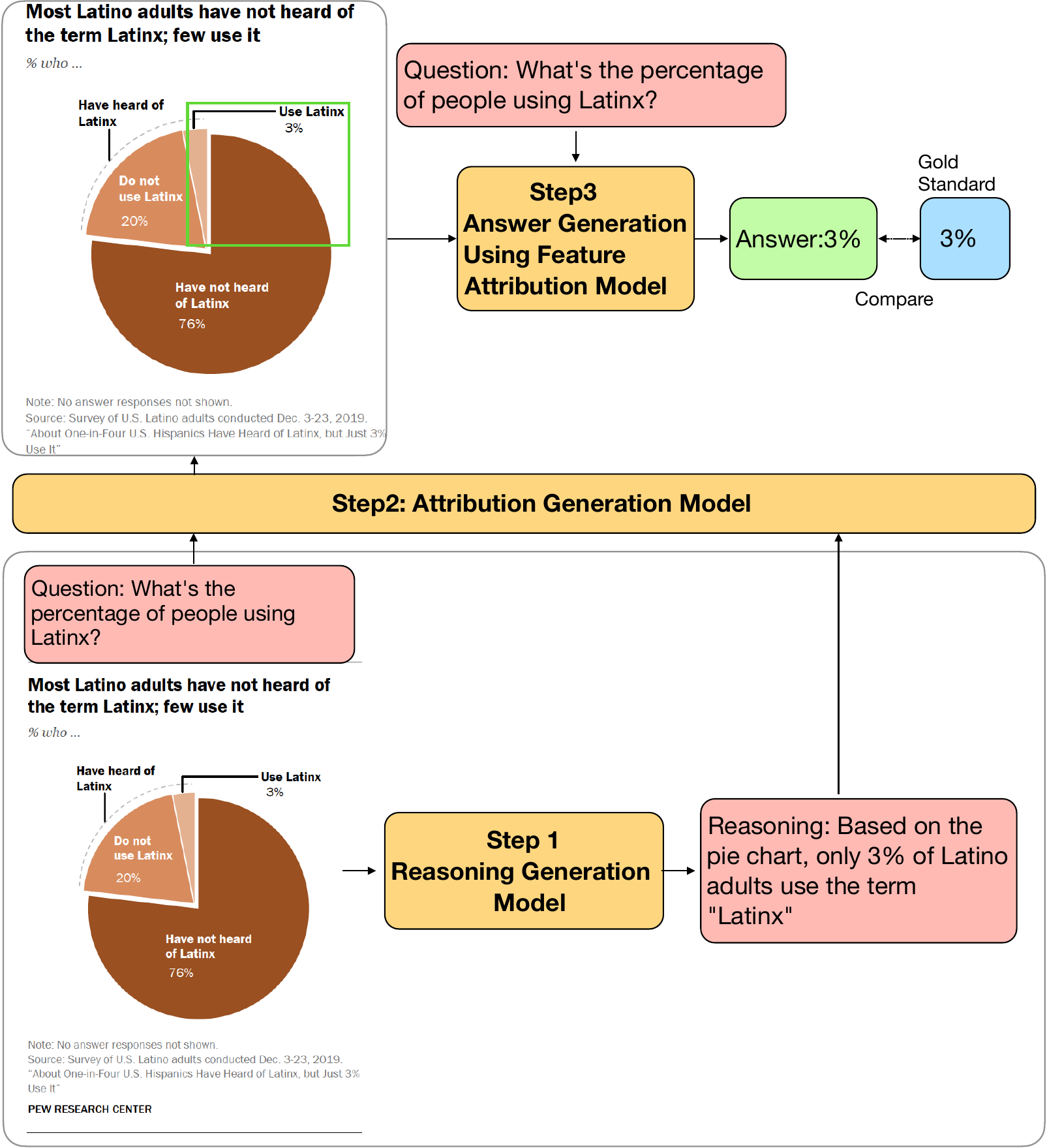}
    \caption{Extension of our proposed framework to pie charts, illustrating the framework's adaptability across different visualization formats.}
    \label{fig:method_overall}
\end{figure}

\noindent \textbf{Generalization \& Scalability:} We extracted 1068 pie charts with their Question, Answer pair from the ChartQA dataset and processed them through three steps: (i) reasoning generation, (ii) attribution generation, and (iii) answer generation using feature attribution as described in Figure \ref{fig:method_overall}. The generated answers, when evaluated to test the attribution system's accuracy, achieve an average BERTScore of $\sim$ 0.9 and average STS of $\sim$0.5. This fully automated approach demonstrates scalability and generalizes effectively across various chart types when provided with charts and questions.

\vspace{-2mm}
\section{Conclusion}
\vspace{-2mm}
In this paper, we presented a novel framework for chart attribution that combines visual and mathematical reasoning capabilities. Our primary contributions include (i) the formalization of chart attribution for mathematical question answering and reasoning tasks, (ii) a systematic data curation strategy that combines MLLM-generated reasoning with human corrections for reliable attribution annotation, and (iii) a framework that utilizes automatic reasoning steps to improve attribution accuracy. While our approach demonstrates significant improvements over baselines, opportunities remain for enhancing reasoning generation, extending support for complex chart types, and integrating with downstream applications. Our framework provides a foundation for building more trustworthy and interpretable AI systems for mathematical reasoning tasks, paving the way for chart-based systems that can better explain their decision-making processes.

\newpage
\section{Limitations}

While our framework demonstrates promising results for chart attribution through automated reasoning, several important limitations and areas for discussion emerge from our study:

\textbf{Restrictive Prompting:} To produce bounding boxes, we use a highly restrictive prompting approach. We instruct the model to generate bounding boxes as a list of coordinate tuples in the format (X1, Y1, X2, Y2). However, research works on restrictive prompting \cite{tam2024letspeakfreelystudy} has found that using overly restrictive prompts can lead to notable decreases in model performance compared to less constrained prompting techniques. Therefore

\textbf{Attribution Task is Challenging for Humans:} As shown in Table 2, the attribution task proved difficult for human annotators. In stages 2 and 3 of the annotation process, the agreement percentages ranged from just 52\% to 64\%. These relatively low levels of agreement underscore the inherent challenge of the attribution task, even for human raters with domain expertise.

\textbf{Reasoning Quality Dependencies}: Our attribution system's performance depends on the quality of generated reasoning steps. While fine-tuning InternLM-XComposer2 may improve reasoning generation, complex mathematical operations, and multi-step calculations still present challenges, potentially affecting attribution accuracy. We discuss failure cases for reasoning in fig \ref{fig:wordcloud}.

\textbf{Computational Requirements}: The sliding window mechanism used for attribution, while effective, requires significant computational resources, especially for high-resolution charts or when processing multiple reasoning steps. This may impact the system's practicality in real-time applications.

\textbf{Human Validation Process}: While our data curation strategy employs human validation to ensure quality, the subjectivity in reasoning annotation and attribution marking can introduce inconsistencies. The inter-annotator agreement scores suggest room for improvement in standardizing the validation process.

\textbf{Model Architecture Limitations}: The current approach relies on layer 16 hidden states of InternLM-XComposer2, which may not capture all relevant features for attribution. Alternative architectural choices or multi-layer approaches could potentially yield better results.

These limitations point to several promising directions for future research, including more robust reasoning generation mechanisms, efficient attribution algorithms, and improved validation methodologies.
\section{Ethics Statement}
We acknowledge several ethical considerations in our development of chart attribution systems. First, we prioritized transparency by openly documenting our methodology, model limitations, and potential biases in both reasoning generation and attribution accuracy. All training data was properly sourced from public datasets with appropriate licensing, and our human annotation process followed fair labor practices, including equitable compensation (\$15/hour) and clear guidelines. While our system aims to improve accessibility and understanding of quantitative information through transparent reasoning steps, we recognize potential risks of misuse, such as the automated generation of misleading chart interpretations. We recommend deploying this technology with appropriate human oversight in high-stakes scenarios and maintaining regular audits for systematic biases. Our goal is to advance chart interpretation capabilities while implementing safeguards that protect against potential misuse and ensure the technology serves its intended purpose of making quantitative information more accessible and understandable to diverse user groups.

\bibliography{anthology,custom}
\bibliographystyle{acl_natbib}
\clearpage
\newpage
\onecolumn
\appendix
\setcounter{section}{0}

\section*{Appendix}\label{sec:appendix}
This section provides additional examples to assist in the understanding and interpretation of the research work presented.

\begin{itemize}
    \item Section~\ref{Appendix_Attribution_Definition}: Attribution Examples
    \item Section~\ref{data_sources}: 
    Dataset Curation
    \item Section~\ref{experiments}: Experiments
\end{itemize}

\section{Attribution Definition \& Examples} 
\label{Appendix_Attribution_Definition}

\begin{figure*}[!ht]
    \centering
    \includegraphics[width=1\textwidth]{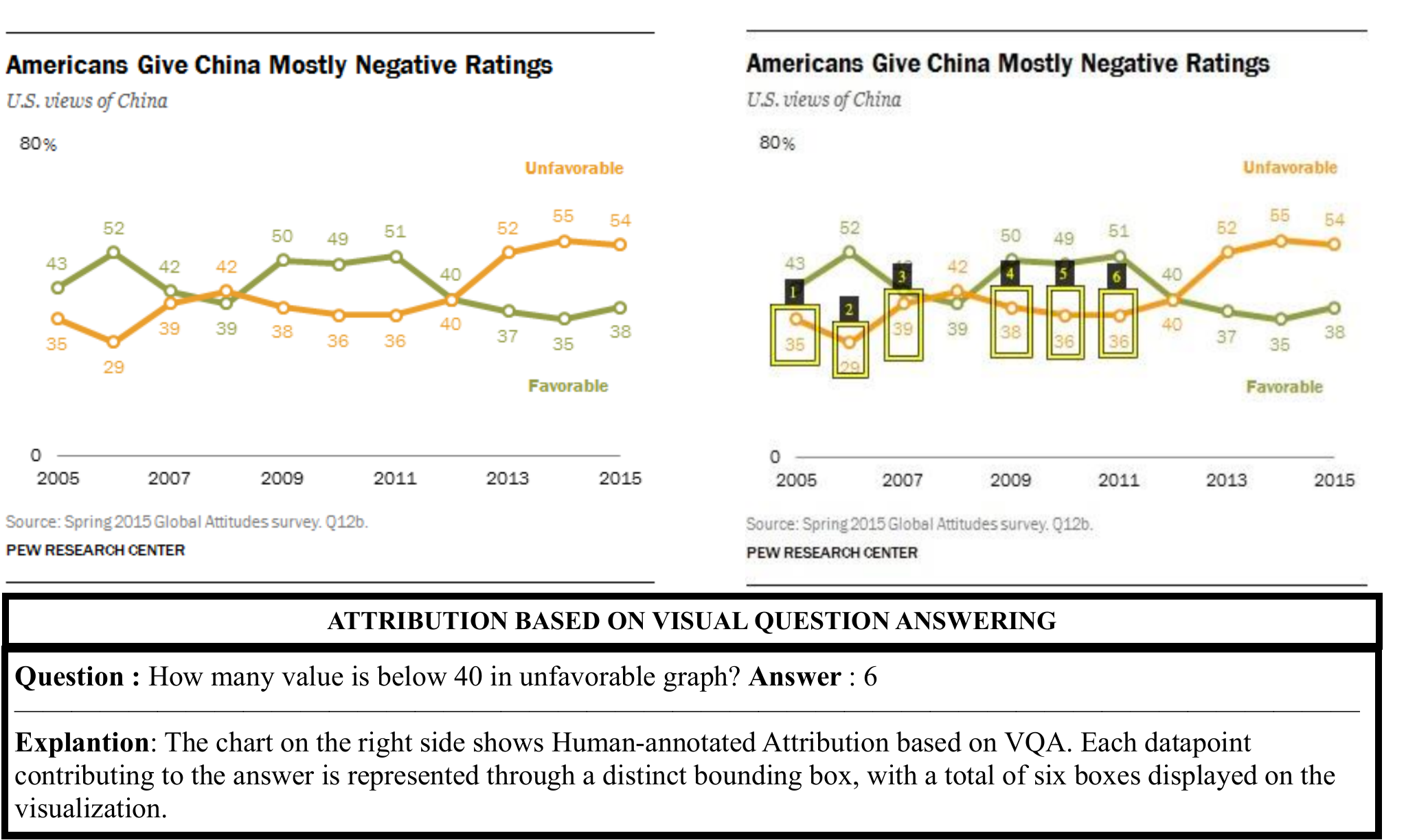}
    \caption{This figure shows attribution based on question answering. Here the bounding boxes clearly identify six data points on the "Unfavorable" line that fall below 40\%, directly supporting the answer to the question "How many values are below 40 in the Unfavorable graph?}
    \label{fig:attribution_vqa}
\end{figure*}

\begin{figure*}[!ht]
    \centering
    \includegraphics[width=1\textwidth]{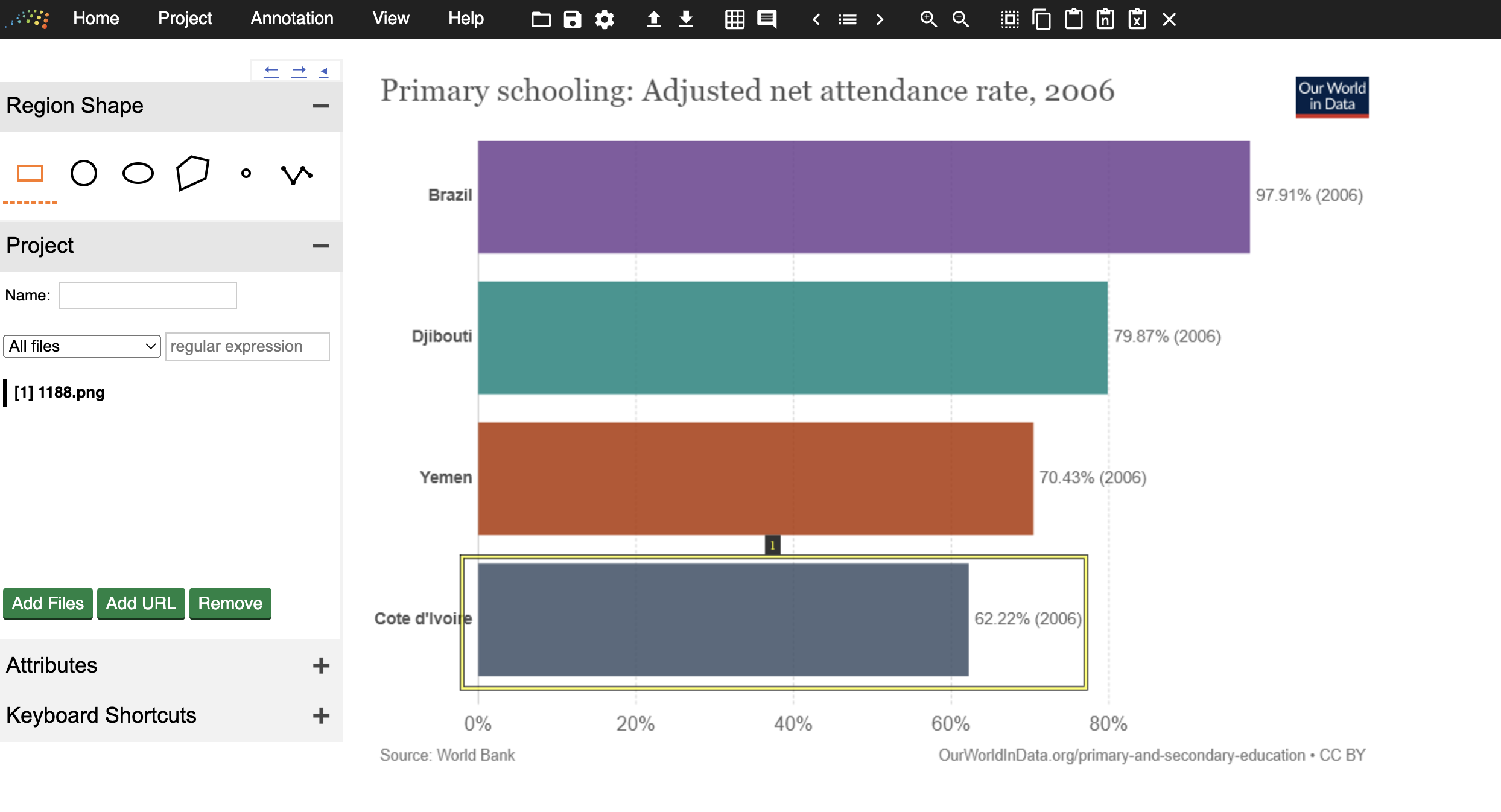}
    \caption{This figure shows attribution annotation platform based on question answering. 
    Annotators are provided question = "What's the value of smallest bar?" and answer="62.22\%". The annotators draw bounding box based on these as represented in the figure.}
    \label{fig:attribution_vqa_ex1}
\end{figure*}

Examples of attribution based on visual question answering is present in fig \ref{fig:attribution_vqa} and \ref{fig:attribution_vqa_ex1}.

\begin{figure*}[!ht]
    \centering
    \includegraphics[width=1\textwidth]{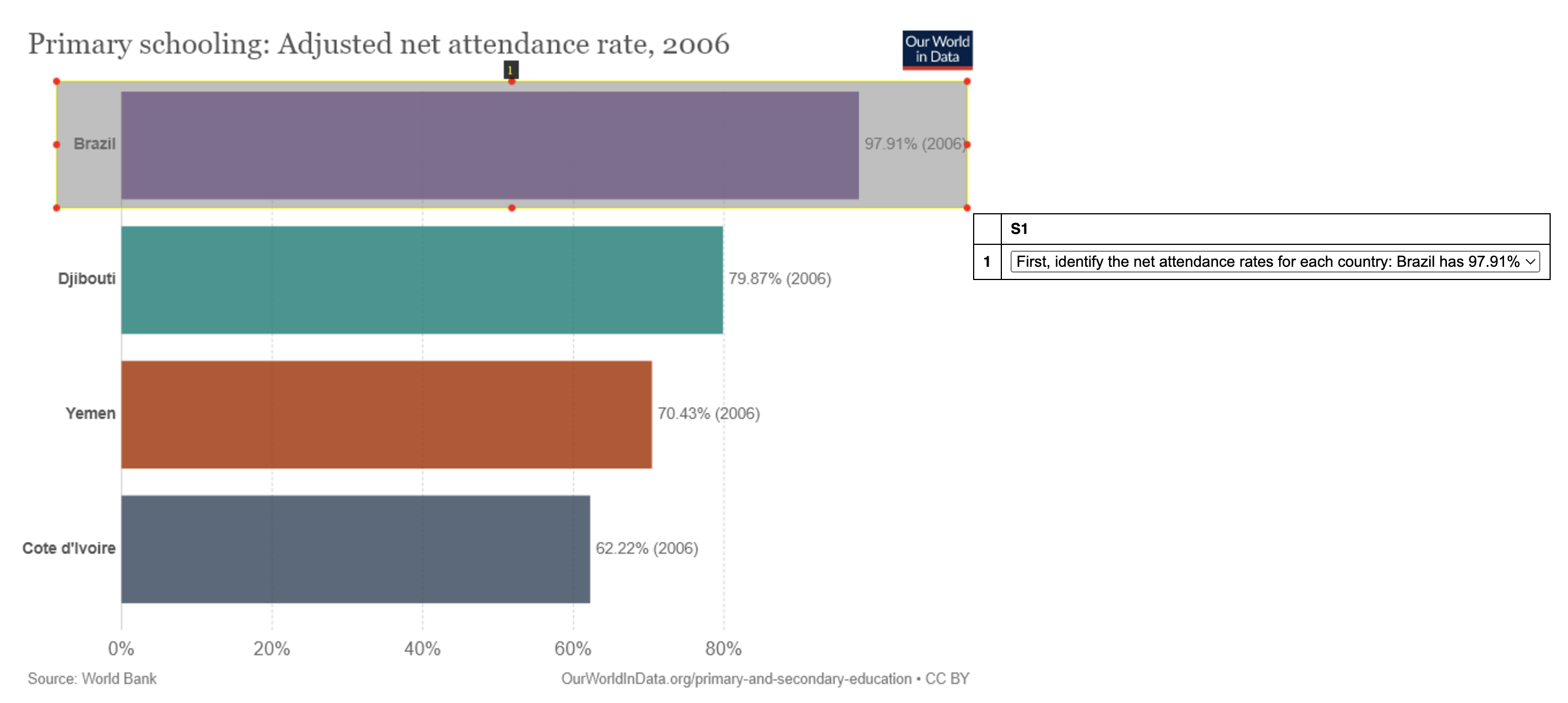}
    \caption{This figure shows how sentence-level reasoning is attributed in our dataset. Annotators are provided with a chart, question = What's the average of Yemen and Brazil?, answer= 84.17, and reasoning = "First, identify the net attendance rates for each country: Brazil has 97.91\%. Yemen has 70.43\%. Next, sum these values: 97.91 + 70.43 = 168.34. Then, divide by the number of countries to find the average: 168.34 / 2 = 84.17". The first reasoning statement "First, identify the net attendance rates for each country: Brazil has 97.91\%." is directly linked to corresponding chart elements, ensuring each step of the mathematical reasoning process is grounded in the chart's components.}
    \label{fig:attribution_vqr_ex2_s1}
\end{figure*}

\begin{figure*}[!ht]
    \centering
    \includegraphics[width=1\textwidth]{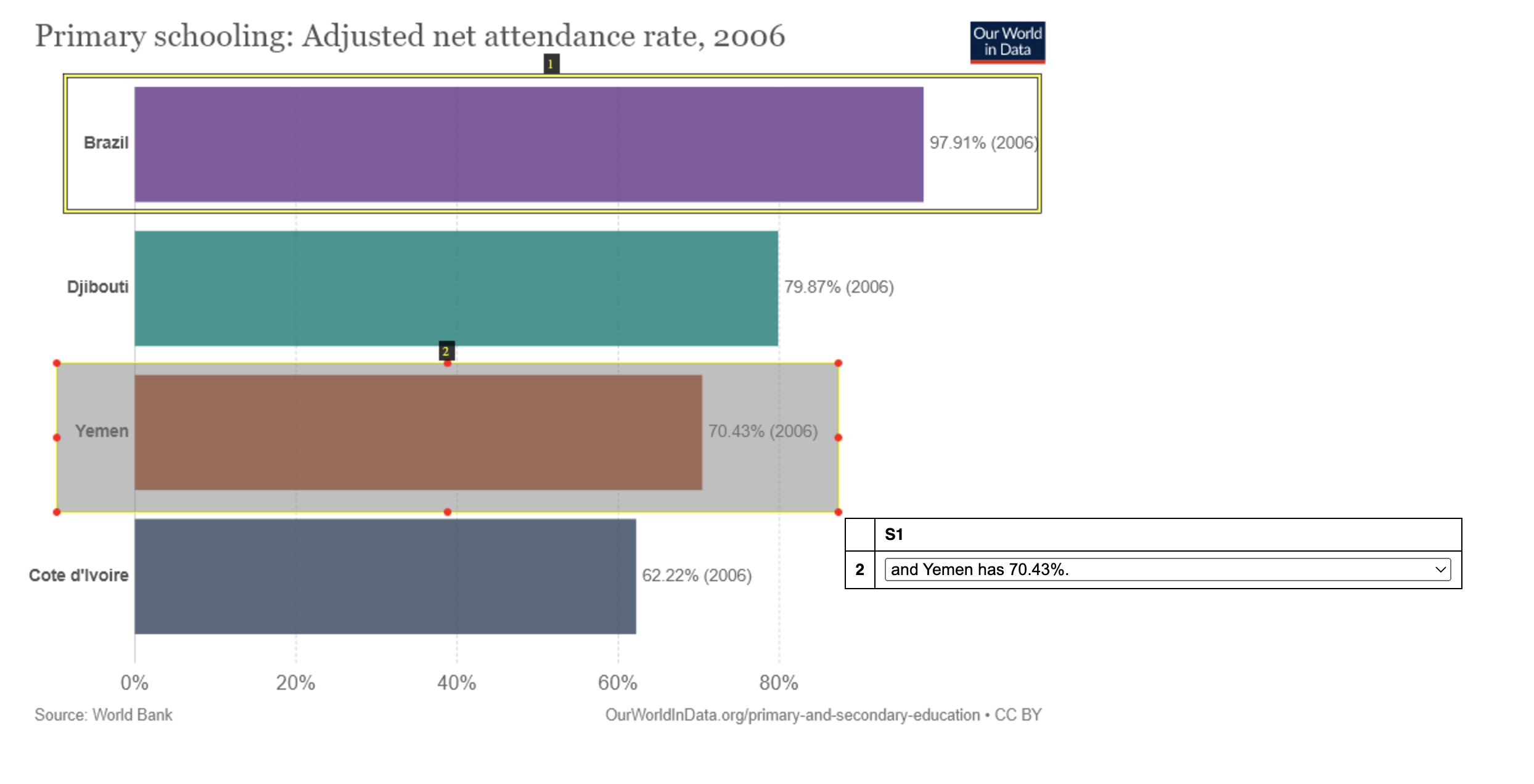}
    \caption{This figure shows how sentence-level reasoning is attributed in our dataset. Annotators are provided with a chart, question = What's the average of Yemen and Brazil?, answer= 84.17, and reasoning = "First, identify the net attendance rates for each country: Brazil has 97.91\%. Yemen has 70.43\%. Next, sum these values: 97.91 + 70.43 = 168.34. Then, divide by the number of countries to find the average: 168.34 / 2 = 84.17". The second reasoning statement "Yemen has 70.43\%" is directly linked to corresponding chart elements, ensuring each step of the mathematical reasoning process is grounded in the chart's components.}
    \label{fig:attribution_vqr_ex2_s2}
\end{figure*}

\begin{figure*}[!ht]
    \centering
    \includegraphics[width=1\textwidth]{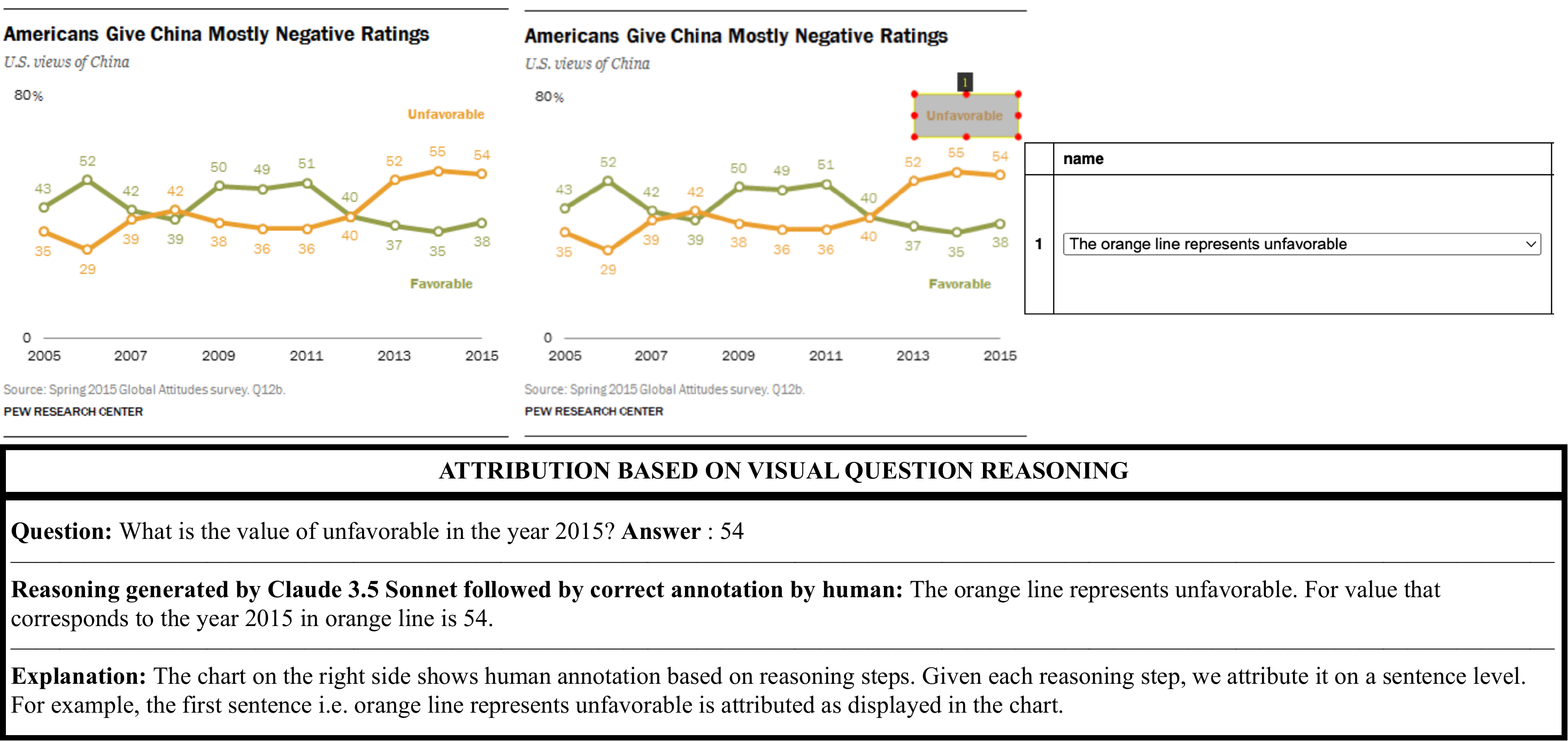}
    \caption{This figure shows how sentence-level reasoning is attributed in our dataset. The first reasoning statement "The orange line represents unfavorable" is directly linked to corresponding chart elements, ensuring each step of the mathematical reasoning process is grounded in the chart's components.}
    \label{fig:attribution_vqr}
\end{figure*}

For Visual Question Reasoning, examples are present in fig \ref{fig:attribution_vqr_ex2_s1}, \ref{fig:attribution_vqr_ex2_s2} and fig \ref{fig:attribution_vqr}.

\vspace{500mm}

\newpage

\newpage

\section{Data Sources and Compilation}

\label{data_sources}

\begin{figure*}[htbp]
\centering
\begin{tikzpicture}[
    scale=0.5,
    transform shape,
    node distance = 1cm and 2cm,
    box/.style = {rectangle, draw, text width=5cm, text centered, minimum height=1cm},
    line/.style = {draw, -latex', blue},
]
\node[box] (A) {ChartQA Dataset};
\node[box, below left=of A] (B) {9.6K human-written questions};
\node[box, below right=of A] (C) {23.1K generated questions};
\node[box, below=of B] (D) {Categorize by chart type};
\node[box, below left=of D] (E) {Line Charts};
\node[box, below=of D] (F) {Bar Charts};
\node[box, below right=of D] (G) {Pie Charts};
\node[box, below=of $(E)!0.5!(F)$] (H) {Random Sampling};
\node[box, below left=of H] (I) {1000 Line Chart QA pairs};
\node[box, below right=of H] (J) {1000 Bar Chart QA pairs};
\node[box, below=of $(I)!0.5!(J)$] (K) {Final Dataset: 2000 QA pairs};

\path[line] (A) -- (B);
\path[line] (A) -- (C);
\path[line] (B) -- (D);
\path[line] (D) -- (E);
\path[line] (D) -- (F);
\path[line] (D) -- (G);
\path[line] (E) -- (H);
\path[line] (F) -- (H);
\path[line] (H) -- (I);
\path[line] (H) -- (J);
\path[line] (I) -- (K);
\path[line] (J) -- (K);
\end{tikzpicture}
\caption{Data Compilation Process Flowchart. From the ChartQA dataset containing both human-written (9.6K) and generated (23.1K) questions, we focus on the human-written subset for quality assurance. We categorize these by chart type (line, bar, and pie charts), then use random sampling to create a balanced final dataset of 2000 QA pairs, comprising 1000 pairs each for line and bar charts.}
\label{fig:data-compilation-flowchart}
\end{figure*}
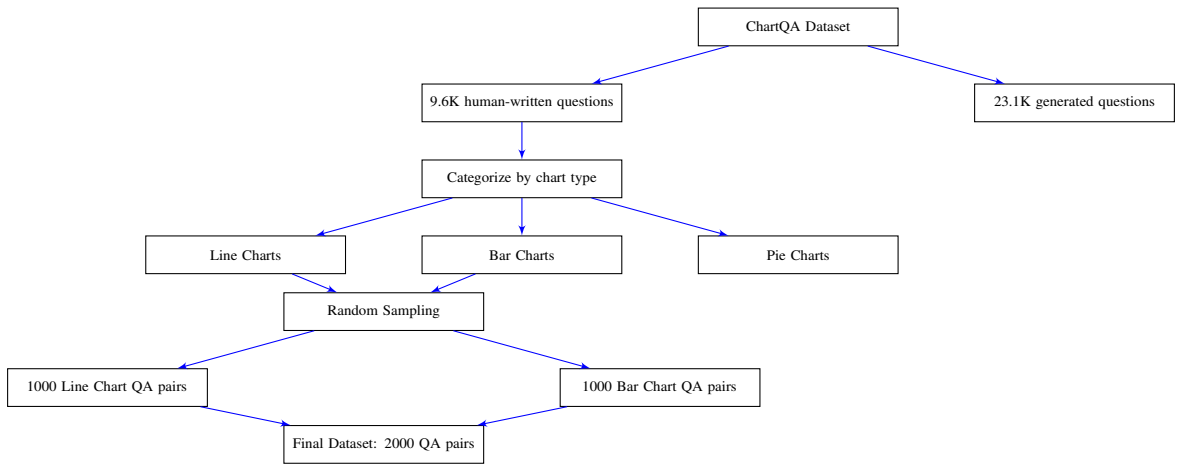

\begin{figure}[!ht]
\includegraphics[width=1\textwidth]{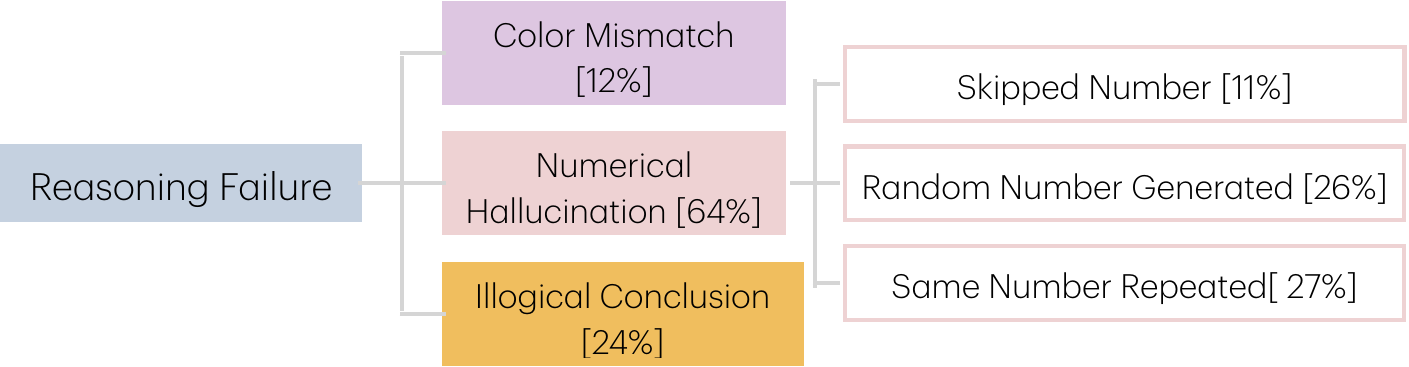}

\caption{Taxonomy of Failure Cases that represents the categories of reasoning failure.}
\label{fig:taxonomy}
\end{figure}

\begin{figure}[!ht]
    \centering
    \includegraphics[width=1\textwidth]{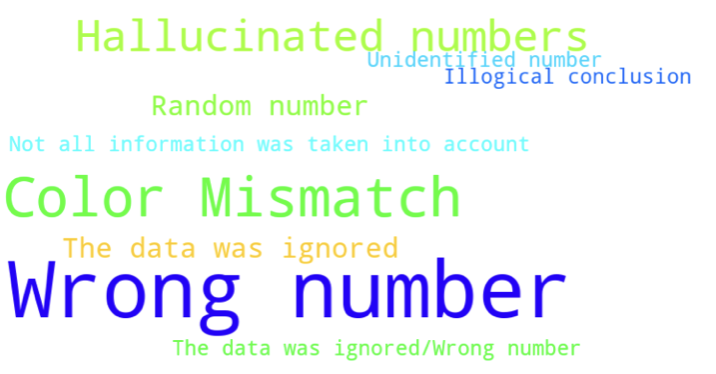}
    \caption{Word cloud representing the annotated labels for reasoning failure. These annotated labels are Hallucinated numbers, Illogical conclusions, Color mismatch, data points ignored, etc.}
    \label{fig:wordcloud}
\end{figure}

\begin{figure*}[!ht]
    \centering
    \includegraphics[width=1\textwidth]{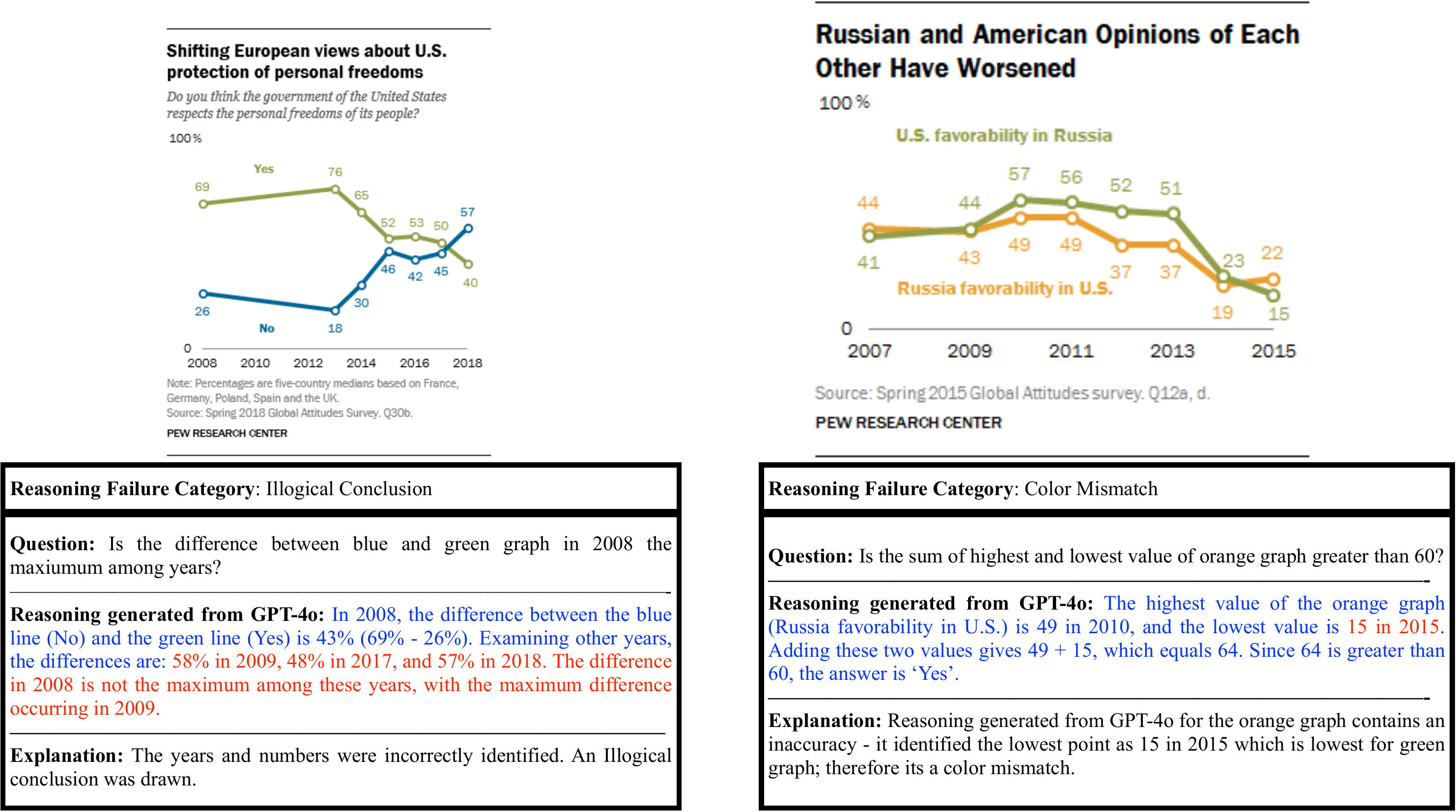}
    \caption{Reasoning failure case examples}
    \label{fig:reasoning_error}
\end{figure*}

\subsection{Data Annotation}
\label{Data_Annotation}

To ensure annotation quality, we conducted initial screening to select mathematically proficient annotators, measured inter-annotator agreement using Kappa score \cite{cohen1960coefficient}, and had authors manually verify a sample of annotations. This semi-automatic approach significantly reduced annotation effort while maintaining high quality through human validation and correction.

The agreement score for stage 1 is based on whether the reasoning is correct and is represented by Kappa score \cite{cohen1960coefficient}. Kappa score is defined as a measure of inter-rater agreement for categorical items, taking into account the agreement occurring by chance.
The Kappa score is defined mathematically as:

$$\kappa = \frac{p_o - p_e}{1 - p_e}$$

where:

$p_o$ is the observed agreement between the two annotators
$p_e$ is the expected agreement by chance

The observed agreement $p_o$ is calculated as:

$$p_o = \frac{a + d}{a + b + c + d}$$

where:
$a$ is the number of cases where both annotators agreed on "yes"
$b$ is the number of cases where the first annotator said "yes" and the second said "no"
$c$ is the number of cases where the first annotator said "no" and the second said "yes"
$d$ is the number of cases where both annotators agreed on "no"

The expected agreement $p_e$ is calculated as:
$$p_e = \frac{(a + b)(a + c) + (c + d)(b + d)}{(a + b + c + d)^2}$$

For stage 2 and stage 3, the Intersection over Union (IOU) score \cite{rezatofighi2019generalized} was calculated. IOU score is a measure of the overlap between the bounding box drawn by annotator 1 and bounding box drawn by annotator 2, defined as the ratio of the area of intersection to the area of union. The IOU score is mathematically defined as:

$$\text{IOU} = \frac{\text{Area of Intersection}}{\text{Area of Union}}$$

where the area of intersection is the overlapping area between the predicted bounding box and the ground truth bounding box, and the area of union is the total area covered by both bounding boxes.

Let's denote the predicted bounding box as $B_p$ and the ground truth bounding box as $B_g$. Then, the IOU score can be calculated as:

$$\text{IOU} = \frac{|B_p \cap B_g|}{|B_p \cup B_g|}$$

Where $|B_p \cap B_g|$ is the area of the intersection and $|B_p \cup B_g|$ is the area of the union.

\newpage
\section{Experiments} 
\label{experiments}

This section contains prompts and additional implementation details.

\subsection{Computing Infrastructure Details}
Our implementation uses PyTorch 2.0 and all experiments were conducted on 4 NVIDIA A100 GPUs with 80GB of memory each. The experiments were run on Amazon Elastic Compute Cloud (Amazon EC2) instances equipped with A100 Tensor Core GPUs and 400 Gbps networking capabilities. The complete experimental pipeline took approximately 100 hours.

\subsection{Prompting Strategies for Attribution} \label{Appendix:prompts}

We experimented with zero-shot and few-shot prompting strategies for both VQA-based and VQR-based attribution.

\begin{zeroshot}

\textcolor{blue}{\textbf{System Prompt}}: You are a helpful assistant that responds in markdown. Help me with my math question.

\textbf{Input Format:}
\begin{itemize}
    \item Chart: [chart\_image], Question: [question\_text], Answer: [answer\_text]
\end{itemize}

\textcolor{blue}{\textbf{User Prompt}}:
Given this chart and the question-answer pair: question = "{question}", answer = "{answer}"; ONLY generate bounding box coordinates in X1, Y1, X2, Y2 format - A list of tuples, each containing (x1, y1, x2, y2) representing the bounding box coordinates without additional text which represents which part of the chart corresponds to the answer.
\end{zeroshot}

\begin{fewshot}

\textcolor{blue}{\textbf{System Prompt}}: You are a helpful assistant that responds in markdown. Help me with my math question.

\textbf{Example 1:}
\begin{example}
Chart: [bar\_chart\_image]
Question: "What was the highest value in 2020?"
Answer: "85 units"

Bounding Box: (120, 45, 140, 230)
\end{example}

\textcolor{blue}{\textbf{User Prompt}}:
Given this chart and the question-answer pair: question = "{question}", answer = "{answer}" and examples; ONLY generate bounding box coordinates in X1, Y1, X2, Y2 format - A list of tuples, each containing (x1, y1, x2, y2) representing the bounding box coordinates without additional text which represents which part of the chart corresponds to the answer.
\end{fewshot}

\begin{figure*}[!ht]
    \centering
    \includegraphics[width=\textwidth]{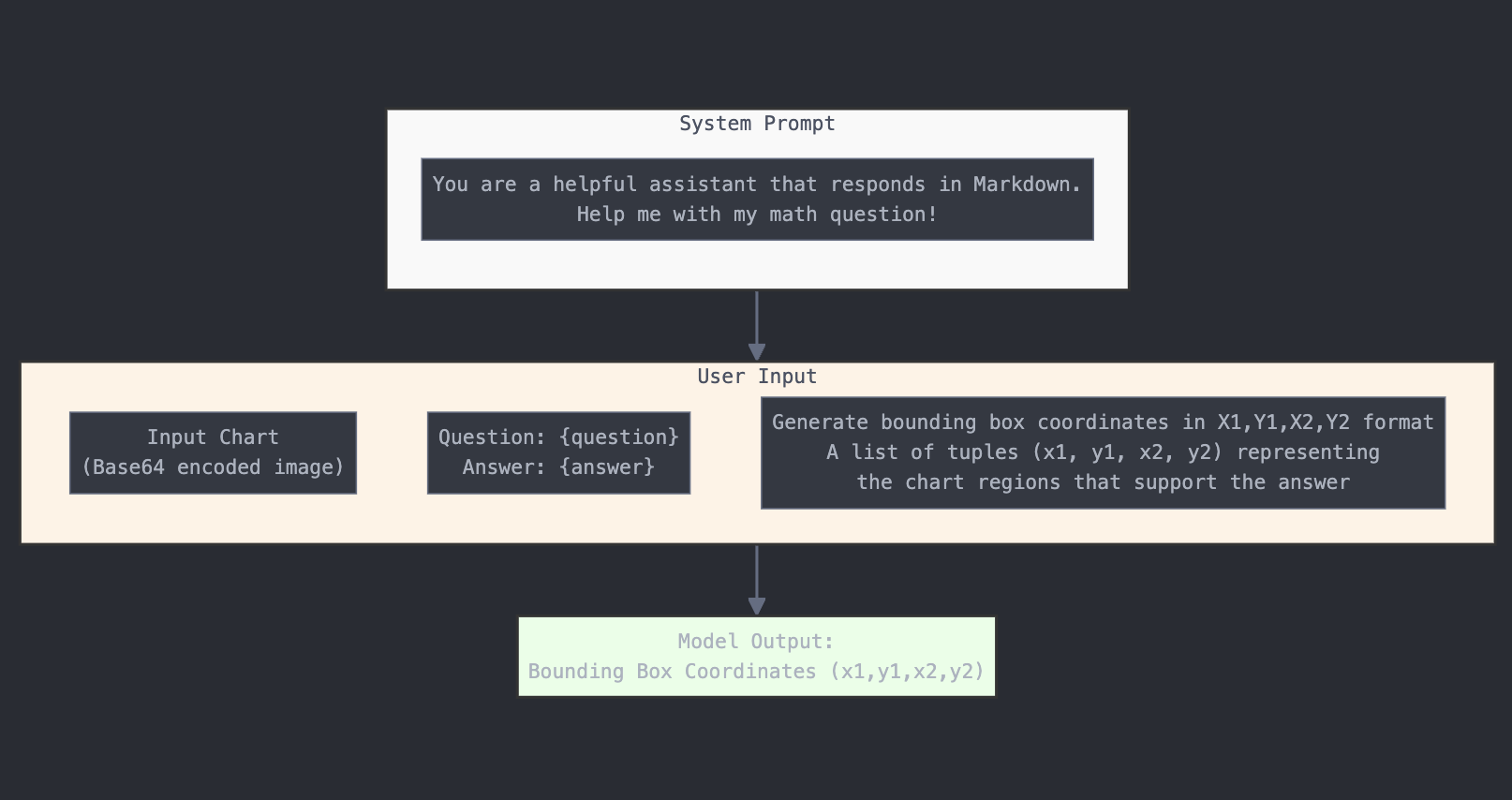}
    \caption{Zero Shot Prompting for Attribution based on VQA task.}
    \label{fig:attribution_VQA_prompt_zeroshot}
\end{figure*}

\begin{figure*}[!ht]
    \centering
    \includegraphics[width=\textwidth]{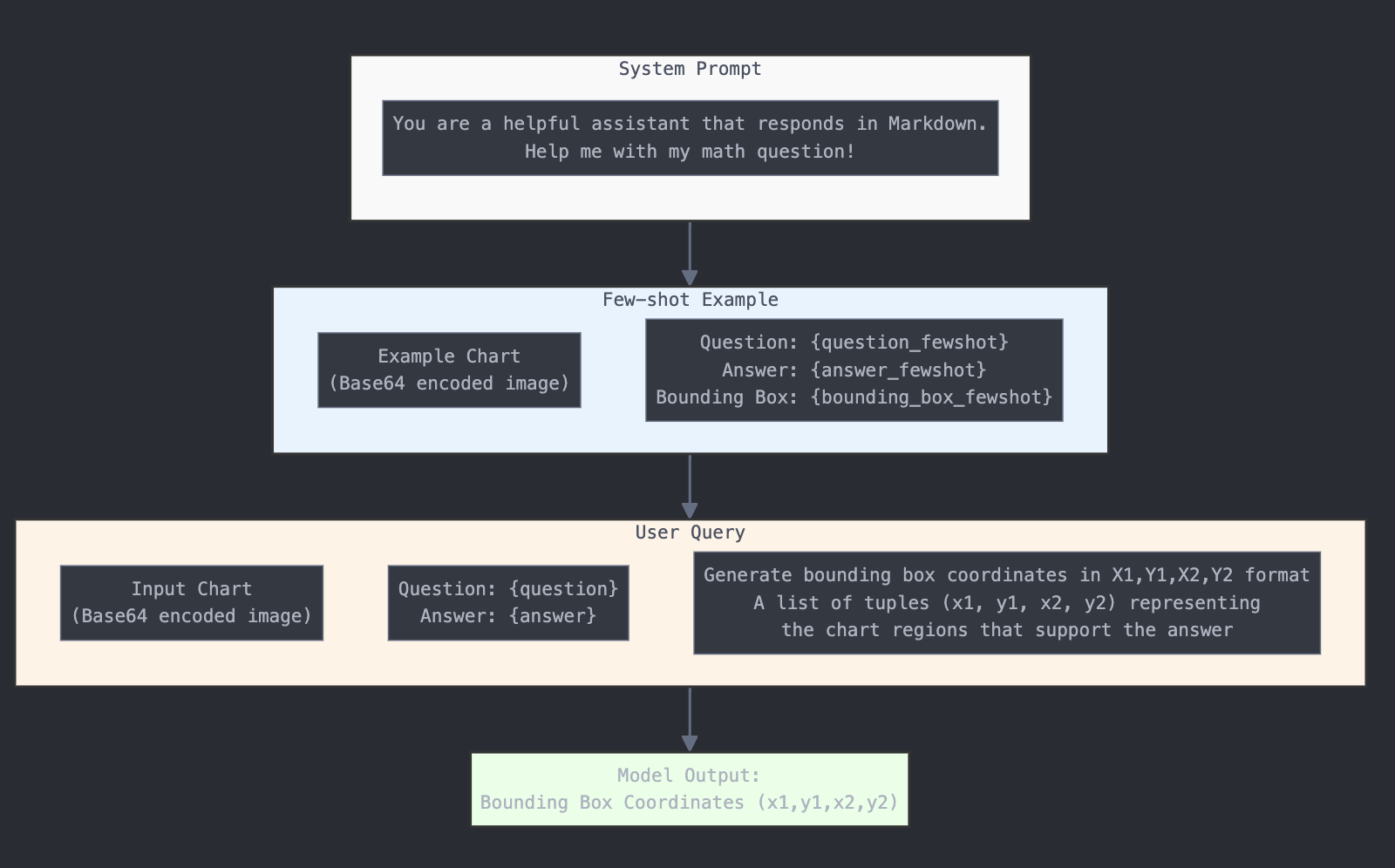}
    \caption{Few Shot Prompting for Attribution based on VQA task.}
    \label{fig:attribution_VQA_prompt_fewshot}
\end{figure*}

\textbf{VQA based Attribution}
For VQA-based Attribution, we used both zero shot and few shot prompting and the prompt is described in figure \ref{fig:attribution_VQA_prompt_zeroshot} and \ref{fig:attribution_VQA_prompt_fewshot}.

\textbf{Attribution based on VQR}

\begin{figure*}[!ht]
    \centering
    \includegraphics[width=\textwidth]{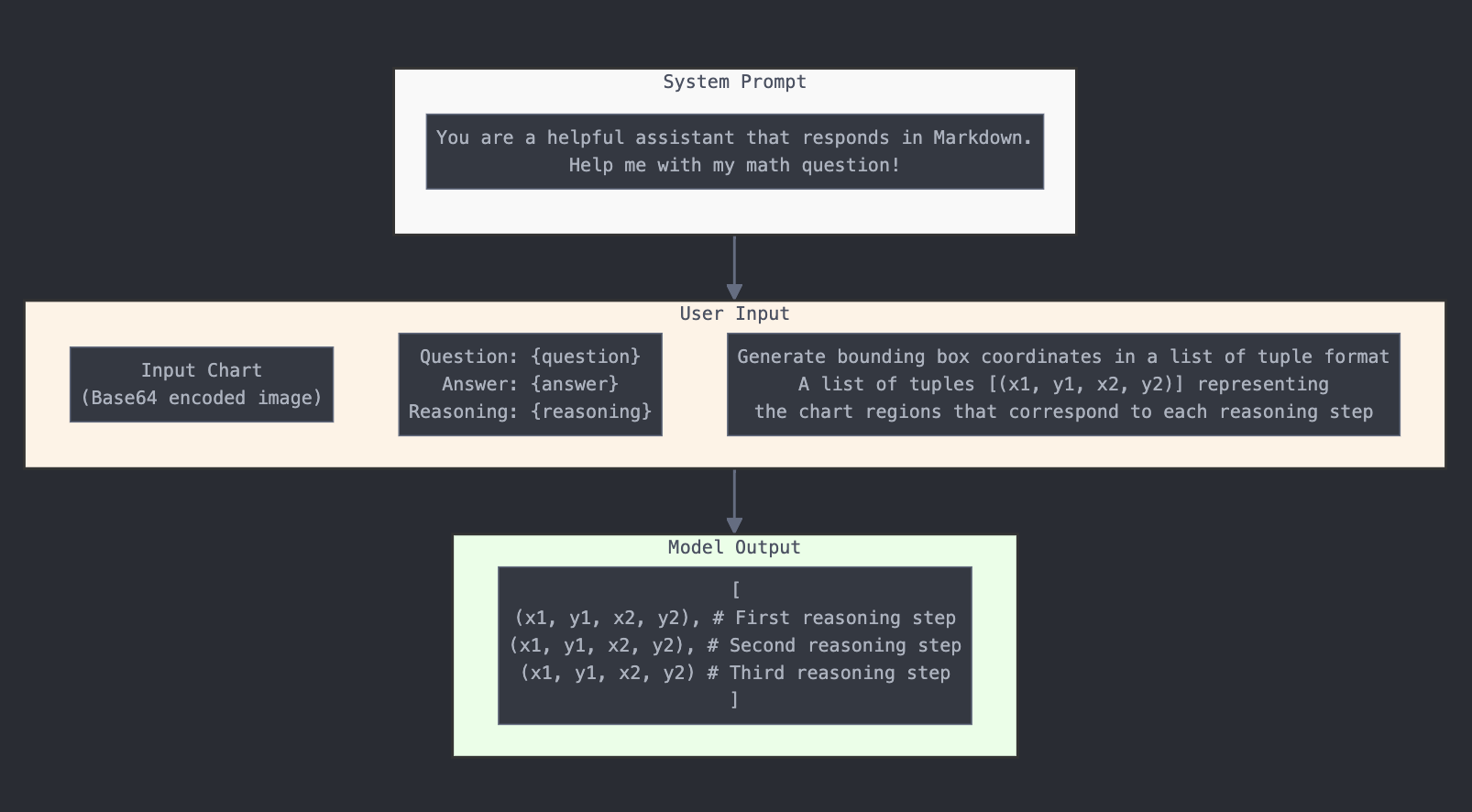}
    \caption{Zero Shot Prompting for Attribution based on VQR task.}
    \label{fig:attribution_VQR_prompt_zeroshot}
\end{figure*}

\begin{figure*}[!ht]
    \centering
    \includegraphics[width=\textwidth]{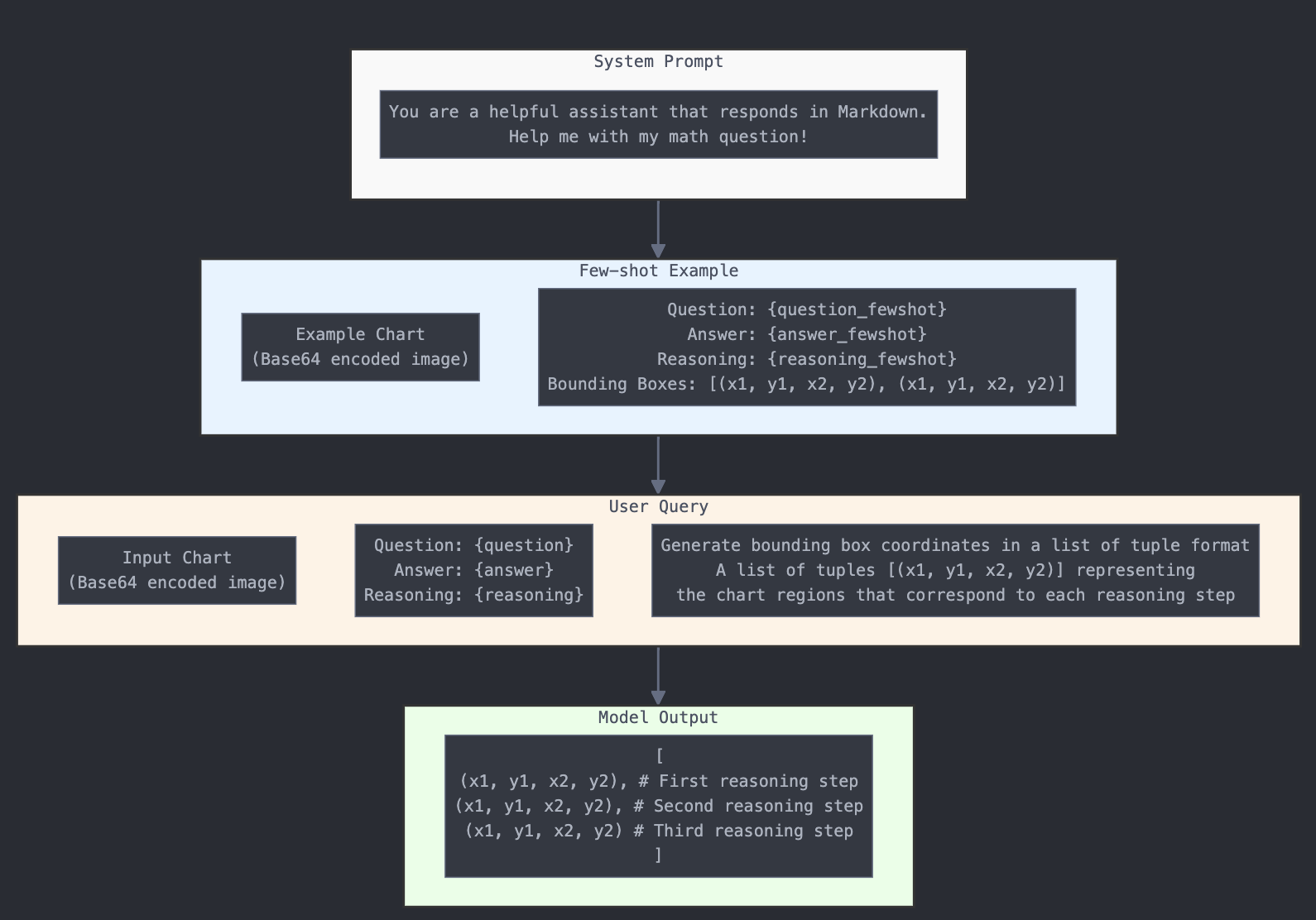}
    \caption{Few Shot Prompting for Attribution based on VQR task.}
    \label{fig:attribution_VQR_prompt_fewshot}
\end{figure*}

For VQR based Attribution, we used both zero shot and few shot prompting and the prompt is described in figure \ref{fig:attribution_VQR_prompt_zeroshot} and \ref{fig:attribution_VQR_prompt_fewshot}.

\subsection{Implementation Details - \name}
\label{Appendix_ATLAS}

The proposed pipeline architecture for chart understanding consists of four integral stages that work in concert to process and analyze chart images with corresponding textual inputs. 

In the first stage, Input Processing, the system handles three primary inputs: the chart image, which serves as the visual input for analysis; the question and answer prompt, which guides the analysis direction. These inputs undergo Base64 encoding for the image and are formatted into a specialized text prompt structure, resulting in encoded inputs suitable for model processing.

The second stage, MLLM Processing, leverages the InternLM-XComposer2 model's multimodal capabilities to process the encoded inputs. This stage extracts Layer 16 Hidden States, which contain rich semantic information from both modalities. The image features are processed as a 35×35 patch grid, while the text features are encoded into 4096-dimensional vectors, enabling comprehensive semantic representation of both visual and textual content. This dual-stream processing ensures that both modalities contribute effectively to the final analysis.

The third stage implements a Sliding Window Attribution mechanism, which is crucial for identifying relevant regions within the chart. This process begins with window generation, where variable-sized windows are created over the image feature space. The system then computes cosine similarity between the text and image features for each window, enabling the identification of regions most pertinent to the textual input. This stage culminates in the selection of the best region, outputting coordinates (i, j, h, w) that specify both the location and dimensions of the most relevant area within the chart.

The final stage focuses on Visualization, transforming the mathematical outputs into interpretable visual representations. This involves coordinate mapping, where the model's internal coordinate space is transformed into image pixel space, followed by bounding box generation that creates visible overlays highlighting the relevant regions identified by the model. This visualization stage is crucial for making the model's decisions interpretable and useful for end users.

The entire pipeline demonstrates flexibility in handling both reasoning-based and answer-based attribution scenarios through the same architectural framework. This unified approach allows for consistent processing while accommodating different types of chart analysis tasks, from simple identification to complex reasoning about chart elements. The system maintains a consistent flow of information through each stage, ensuring that the final output effectively bridges the gap between the visual elements of the chart and the textual understanding required for comprehensive chart analysis.

\begin{figure*}[!ht]
    \centering
    \includegraphics[width=0.5\textwidth]{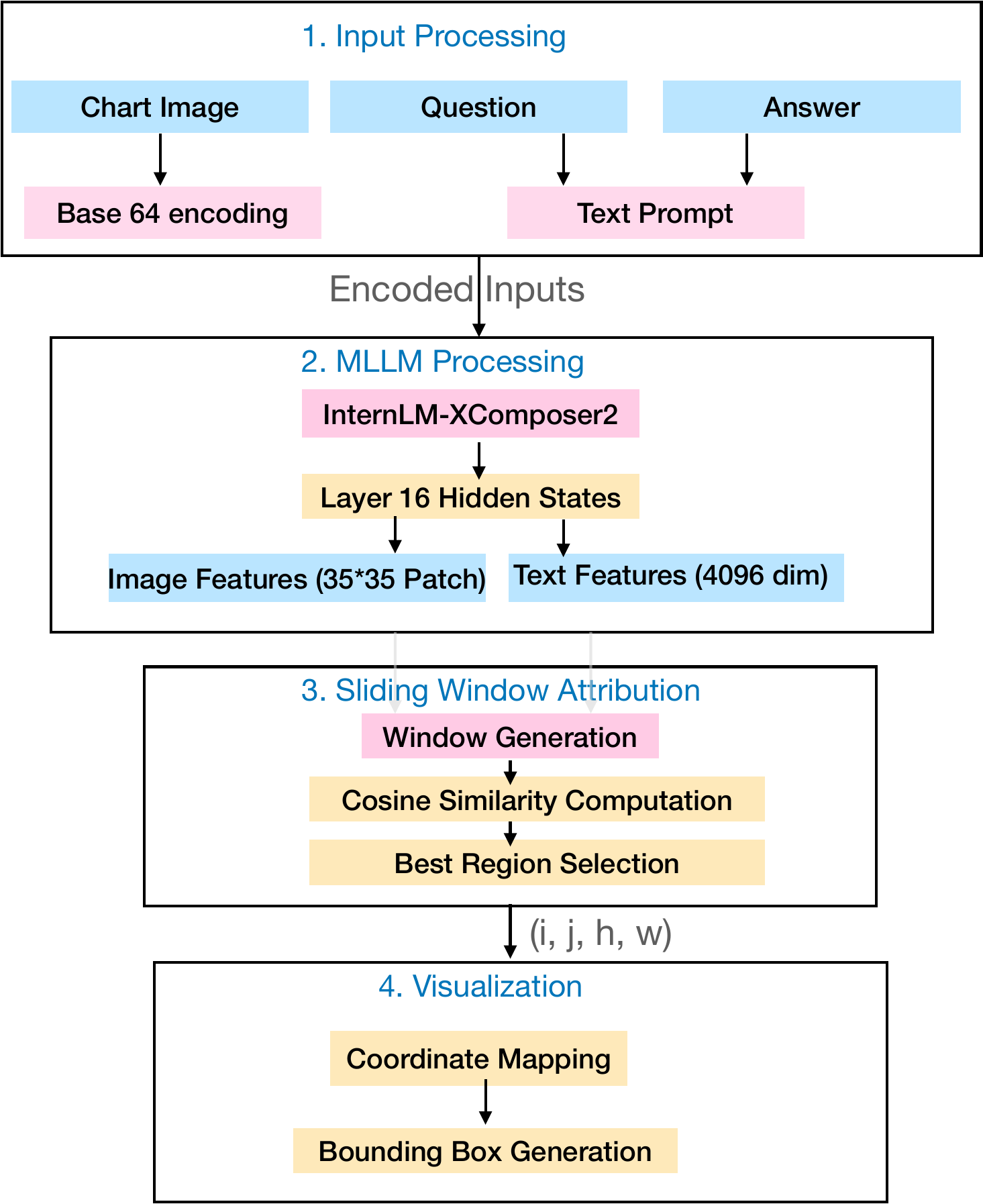}
    \caption{The pipeline architecture for chart understanding with InternLM-XComposer2 illustrates a four-stage process that bridges visual and textual modalities in chart analysis. The system progresses through Input Processing (encoding of chart images and text), MLLM Processing (multimodal feature extraction), Sliding Window Attribution (region identification), and Visualization (interpretable output generation), enabling comprehensive chart understanding through a unified architectural framework. This architecture supports answer-based attribution.}
    \label{fig: InternLMQA}
\end{figure*}

\begin{figure*}[!ht]
    \centering
    \includegraphics[width=0.5\textwidth]{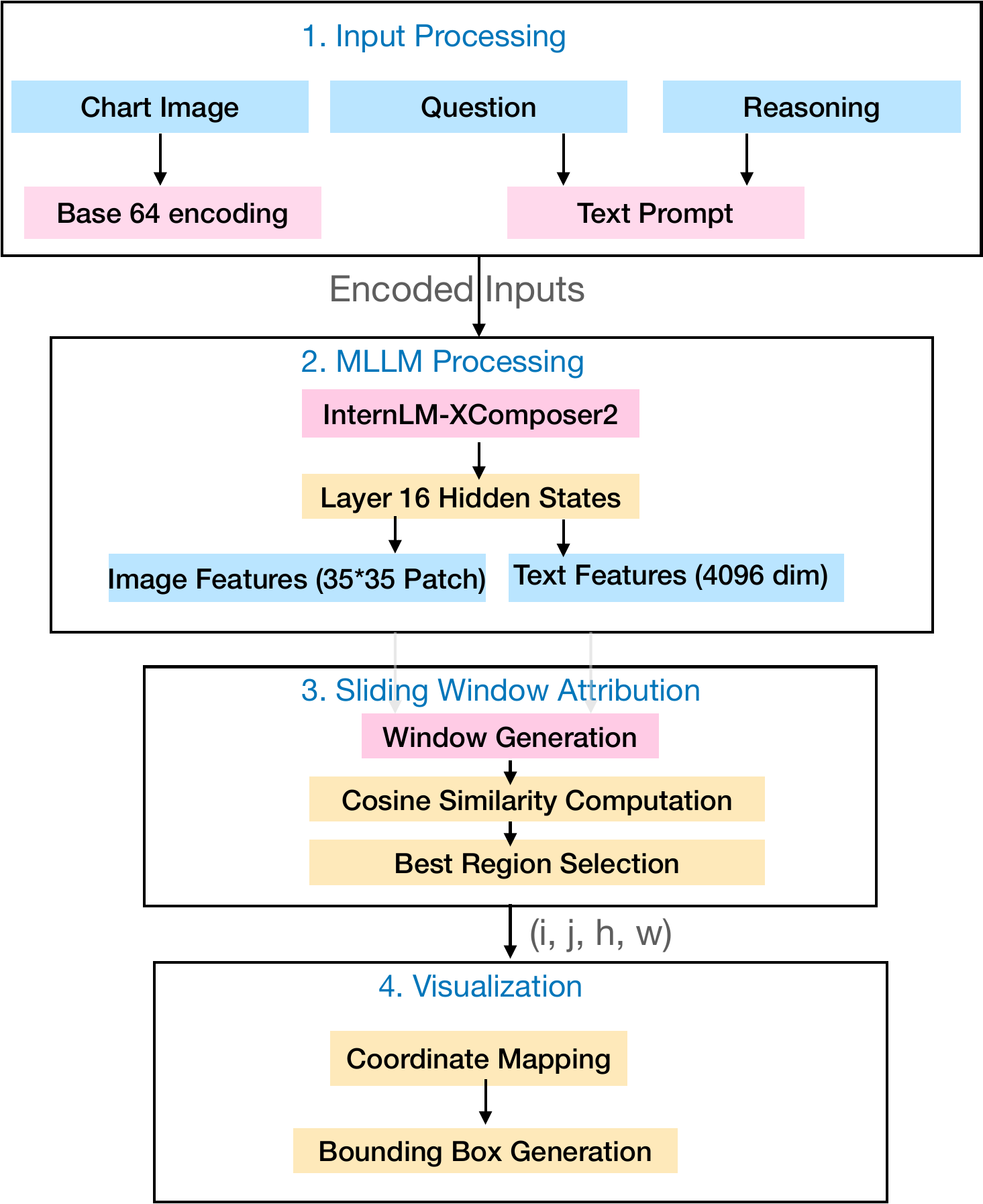}
    \caption{The pipeline architecture for chart understanding with InternLM-XComposer2 illustrates a four-stage process that bridges visual and textual modalities in chart analysis. The system progresses through Input Processing (encoding of chart images and text), MLLM Processing (multimodal feature extraction), Sliding Window Attribution (region identification), and Visualization (interpretable output generation), enabling comprehensive chart understanding through a unified architectural framework. This architecture supports reasoning based attribution.}
    \label{fig: InternLMQAR}
\end{figure*}

Figure \ref{fig: InternLMQA} and \ref{fig: InternLMQAR} represents VQA based and VQR based attribution details respectively.

\begin{figure*}[t]
    \centering
    \includegraphics[width=1\textwidth]{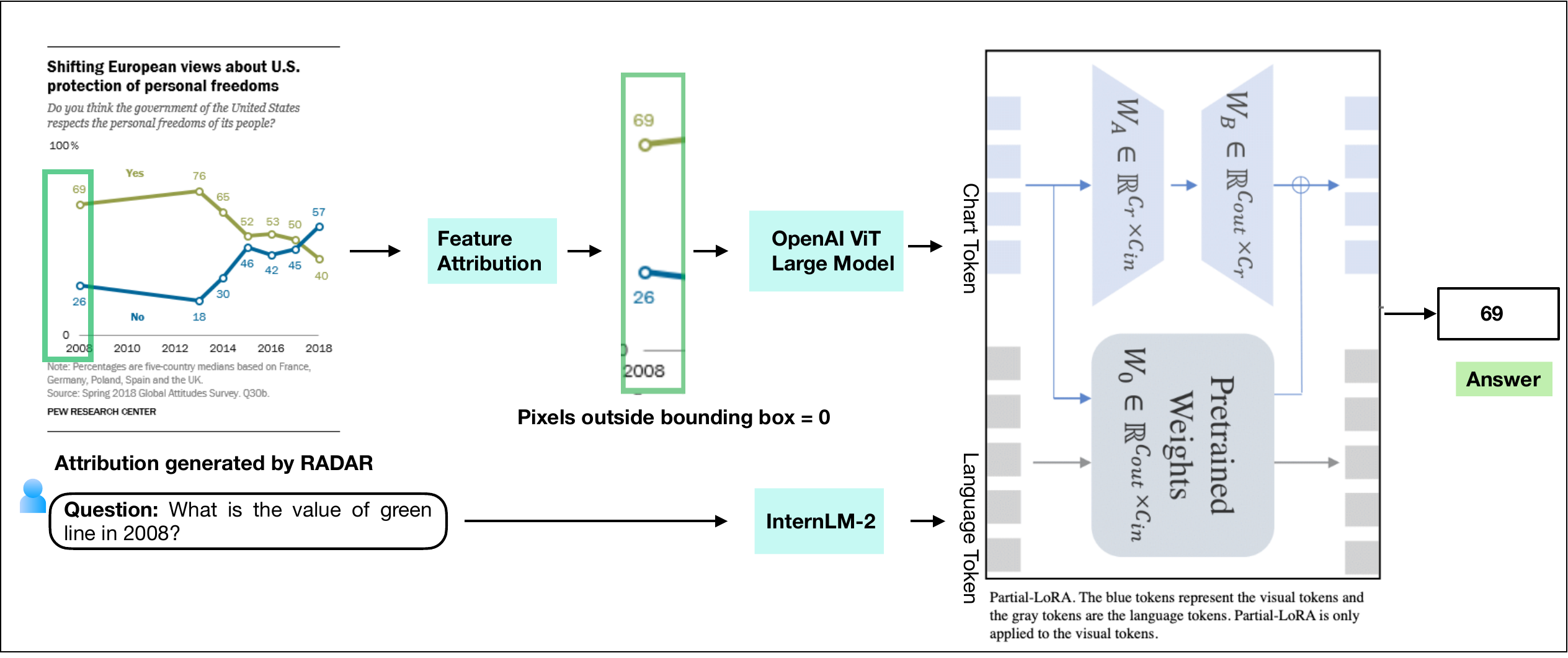}
    \caption{Automatic Attribution Evaluation: Attribution generated by \name\ is passed through feature attribution that sets the pixels outside the bounding box to zero. Feature Attributed Charts and Question is passed as input through the framework to collect answers. The generated answer is then matched with the ground truth.}
    \label{fig:Reverse_Evaluation}
\end{figure*}

\subsection{Automatic Reasoning Evaluation} \label{Appendix:Reasoning_Evaluation}

Bertscore \cite{Zhang2019BERTScoreET} computes the similarity between two sentences using contextual embeddings from BERT.

Semantic Textual Similarity (STS) \cite{agirre-etal-2012-semeval} measures how similar two pieces of text are in terms of their meaning, regardless of their exact wording.

$ \text{STS}(s_1, s_2) = \cos(\vec{v_1}, \vec{v_2}) = \frac{\vec{v_1} \cdot \vec{v_2}}{||\vec{v_1}|| \, ||\vec{v_2}||} $

Where $\vec{v_1}$ is the vector representation (embedding) of the reasoning generated from \name\ and
$\vec{v_2}$ is the vector representation (embedding) of the reasoning that is human annotated. $s_1$, and $s_2$ represent the reasoning generated from \name\ and human annotated ones.
This equation measures the semantic similarity between \name-generated reasoning and human-annotated reasoning by converting both reasonings into vector representations and computing their cosine similarity using dot product and vector magnitudes.

After all the scores were calculated, we took an average and the scores are reported in table \ref{tab:reasoning_scores}.

\end{document}